\documentclass[pdflatex,sn-mathphys-num]{sn-jnl}

\usepackage{graphicx}%
\usepackage{multirow}%
\usepackage{amsmath,amssymb,amsfonts}%
\usepackage{amsthm}%
\usepackage{mathrsfs}%
\usepackage[title]{appendix}%
\usepackage{xcolor}%
\usepackage{textcomp}%
\usepackage{manyfoot}%
\usepackage{booktabs}%
\usepackage{algorithm}%
\usepackage{algorithmicx}%
\usepackage{algpseudocode}%
\usepackage{listings}%
\usepackage{pdfpages}

\usepackage[version=4]{mhchem}
\newcommand{\pKa}{\ensuremath{\text{p}K_{\mathrm{a}}}}
\usepackage{comment}
\usepackage[T1]{fontenc}

\theoremstyle{thmstyleone}%
\theoremstyle{thmstyletwo}%

\theoremstyle{thmstylethree}%

\begin{document}

\title{Electron flow matching for generative reaction mechanism prediction obeying conservation laws}


\author[1,2]{\fnm{Joonyoung F.} \sur{Joung}}\email{jjoung@mit.edu}
\equalcont{These authors contributed equally to this work.}
\author[1]{\fnm{Mun Hong} \sur{Fong}}\email{fong410@mit.edu}
\equalcont{These authors contributed equally to this work.}

\author[1]{\fnm{Nicholas} 
\sur{Casetti}}\email{ncasetti@mit.edu}

\author[1]{\fnm{Jordan P.} \sur{Liles}}\email{jliles@mit.edu}
\author[3]{\fnm{Ne S.} \sur{Dassanayake}}\email{nedassa@mit.edu}

\author*[1,4]{\fnm{Connor W.} \sur{Coley}}\email{ccoley@mit.edu}

\affil*[1]{\orgdiv{Department of Chemical Engineering}, \orgname{Massachusetts Institute of Technology}, \orgaddress{\street{77 Massachusetts Ave.}, \city{Cambridge}, \postcode{02139}, \state{Massachusetts}, \country{United States}}}

\affil[2]{\orgdiv{Department of Chemistry}, \orgname{Kookmin University}, \orgaddress{\street{77 Jeongneung-ro, Seongbuk-gu}, \city{Seoul}, \postcode{02707}, \country{Republic of Korea}}}

\affil[3]{\orgdiv{Department of Chemistry}, \orgname{Massachusetts Institute of Technology}, \orgaddress{\street{77 Massachusetts Ave.}, \city{Cambridge}, \postcode{02139}, \state{Massachusetts}, \country{United States}}}

\affil[4]{\orgdiv{Department of Electrical Engineering and Computer Science}, \orgname{Massachusetts Institute of Technology}, \orgaddress{\street{77 Massachusetts Ave.}, \city{Cambridge}, \postcode{02139}, \state{Massachusetts}, \country{United States}}}


\abstract{Central to our understanding of chemical reactivity is the principle of mass conservation, which is fundamental for ensuring physical consistency, balancing equations, and guiding reaction design. However, data-driven computational models for tasks such as reaction product prediction rarely abide by this most basic constraint. 
In this work, we recast the problem of reaction prediction as a problem of electron redistribution using the modern deep generative framework of flow matching.  
Our model, FlowER, overcomes limitations inherent in previous approaches by enforcing exact mass conservation, thereby resolving hallucinatory failure modes, recovering mechanistic reaction sequences for unseen substrate scaffolds, and generalizing effectively to out-of-domain reaction classes with extremely data-efficient fine-tuning. 
FlowER additionally enables estimation of thermodynamic or kinetic feasibility
and manifests a degree of chemical intuition in reaction prediction tasks. This inherently interpretable framework represents a significant step in bridging the gap between predictive accuracy and mechanistic understanding in data-driven reaction outcome prediction.}


\keywords{machine learning, generative AI, reaction mechanism, computer-aided synthesis planning}

\maketitle

\section{Introduction}\label{sec1}
Mass conservation is a fundamental principle in chemistry, servicing as a critical constraint for accurately modeling chemical reactions. Postulated by Antoine Lavoisier in the eighteenth century, it asserts that the total mass of reactants equals the total mass of products, forming the basis for stoichiometry and chemical equation balancing. Despite its simplicity and essentiality, many machine learning models trained on chemical reaction data do not inherently enforce mass conservation.  
In this work, we introduce a new modeling formulation for reaction outcome prediction that achieves exact conservation 
by modeling chemical reactivity as a generative and probabilistic process of electron redistribution. 
 
The task of reaction outcome prediction has become a popular target for supervised machine learning \cite{schwaller2022machine, tu2023predictive}.
While chemists typically conceptualize, visualize, and communicate understanding of chemical reactions through mechanistic arrow-pushing diagrams, most data-driven models bypass this formalism and focus solely on predicting the major product in an end-to-end manner. 
Common approaches include predicting bond-breaking and formation as edits to molecular graphs \cite{do2019graph, jin2017predicting, coley2019graph}, inferring pseudo-mechanisms from overall reactions \cite{bradshaw2018generative, bi2021non}, or translating reactant(s) to product(s) in the language of SMILES strings \cite{schwaller2019molecular, tetko2020state, tu2022permutation}.
These ``black-box'' methods are unable to explain \emph{why} a certain product is predicted and can perform unpredictably on extrapolative, out-of-distribution settings \cite{bradshaw2025challenging}.

Limited efforts have been made to train predictive models that understand transformations in synthetic organic chemistry in terms of individual mechanistic steps. Pioneering work from Baldi has produced a range of approaches trained on textbook reactions that rank possible electron flows between donors and acceptors \cite{chen2009no, kayala2011learning, kayala2012reactionpredictor, fooshee2018deep, tavakoli2024ai}. In our own prior work, we have adapted methods from end-to-end reaction prediction to operate at the level of elementary steps by retraining on mechanistic reaction data \cite{joung2024reproducing}.  
However, deep learning architectures are 
prone to the hallucination of atoms or electrons,  resembling ``alchemy'', and---excepting work from Baldi---fail to adhere to fundamental physical laws \cite{wang2022theory, schwaller2022machine, joung2024reproducing}.  
Such limitations expose a critical gap in current machine learning approaches to reaction prediction. 

Generative modeling has become a robust framework for learning complex data distributions and offers a path toward mitigating these limitations. 
Among generative modeling approaches, score-based \cite{song2020score} and denoising diffusion probabilistic models \cite{ho2020denoising} have emerged as particularly popular frameworks.
Both approaches have demonstrated success in molecular applications such as \emph{de novo} protein design \cite{watson2023novo, yim2023fast, ingraham2023illuminating, krishna2024generalized}, small molecule generation \cite{hoogeboom2022equivariant, xu2023geometric, huang2023mdm}, retrosynthesis \cite{igashov2023retrobridge, wang2023retrodiff, laabid2024alignment} and transition state prediction \cite{kim2024diffusion, duan2023accurate}. 
Flow matching generalizes diffusion-based approaches, offers faster inference, and enables arbitrary prior distribution sampling while maintaining or even improving sample quality over diffusion models \cite{tong2023conditional, lipman2022flow, liu2022flow}.

\begin{figure*}[htbp]
\centering
\includegraphics[width=\textwidth]{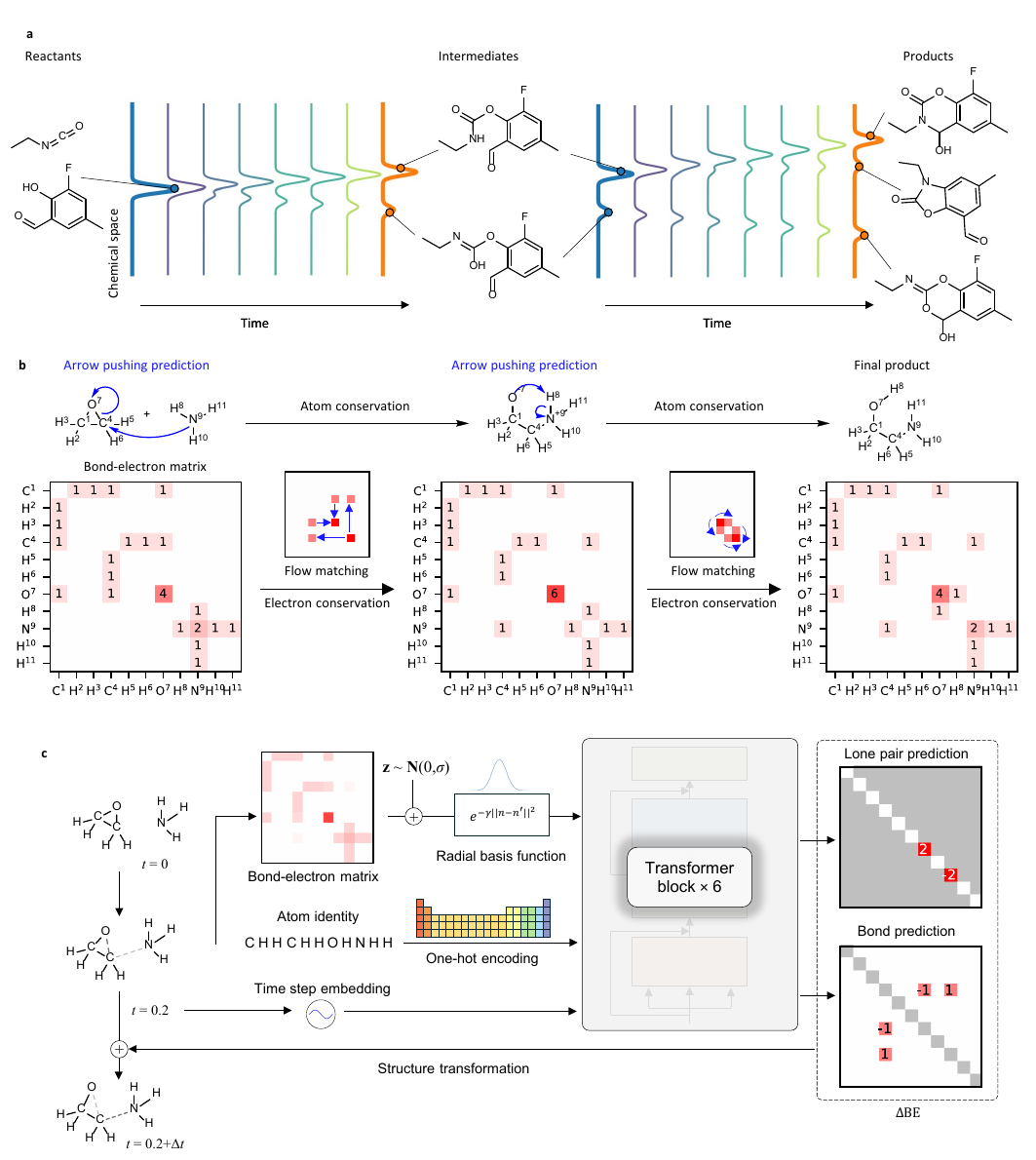}
\caption{
\textbf{a}. Representation of flow in a hypothetical 1D chemical space. Starting from the initial reactant electron configuration on the far left, the reaction progresses probabilistically through intermediate states to produce a distribution over products on the far right. Each molecule occupies a position in a 1D chemical space, representing its state during the reaction. The transition between states represents the flow of the reaction, wherein the electron configuration is gradually redistributed, ultimately transforming the reactant into the product.
\textbf{b}. To enforce strict conservation, FlowER formalizes a chemical reaction as the \emph{redistribution} of valence electrons between atoms present among the reacting species. The state of a system with fixed atomic identities is represented by a bond-electron (BE) matrix as championed by Ugi in the 1970s \cite{10.1007/BFb0051317, ugi1993computer}. The electron redistribution process is learned through flow matching, and electrons are conserved throughout.
\textbf{c}. FlowER takes in a ``transient" state at any given timepoint as input. The molecular structure is represented by a BE matrix and a matrix of atom features. This information is processed by a series of transformer blocks employing a multi-head attention mechanism to finally predict the changes in lone pair and bond electrons, the sum of which is constrained to be zero. 
}\label{fig1}
\end{figure*}

This work introduces   \textbf{flow} matching for \textbf{e}lectron \textbf{r}edistribution (FlowER), a model that represents chemical reactions as a generative process of electron redistribution.
This perspective 
captures the physical principles of conservation, conceptually aligns with arrow-pushing formalisms, and thus achieves useful interpretive clarity. By employing the generative paradigm of flow matching, we capture the probabilistic nature of chemical reactions where multiple outcomes are reached through branching mechanistic networks evolving in time (Fig.~\ref{fig1}a).  
Conservation of all atoms and electrons is accomplished by explicitly tracking electron movement to ensure physically realistic predictions 
using the Bond-Electron (BE) matrix framework championed by Ugi \cite{10.1007/BFb0051317, ugi1993computer}, where atomic identities and their electron configurations are encoded as a compact representation of the reaction system (Fig.~\ref{fig1}b). 
FlowER’s primary task is to predict changes in the BE matrix that describe how electrons redistribute from reactants (the prior distribution) to products (the target distribution). 
Parameterizating predictions with deep neural networks enables
generalization to electron redistribution across a wide variety of chemical reactions.

We demonstrate that FlowER's mass-conserving flow matching strategy provides a physically grounded understanding of chemical reactions and achieves impressive empirical performance on the task of reaction outcome prediction. 
The model excels in recovering complete mechanistic sequences with strict mass conservation, learning fundamental chemical principles directly from data that connect to expert intuition. Conservation further enables downstream thermodynamic evaluations of reaction feasibility.
Finally, we demonstrate that FlowER achieves impressive fine-tuning performance on unseen reaction classes with only 32 reaction examples, demonstrating adaptability with unprecedented sample efficiency.

\section{Results}
\subsection{FlowER is a generative flow matching model for predicting electron redistribution}\label{sec_FlowER}

FlowER models the movement of electrons as a continuous process. 
The state of a reacting system is defined by a fixed set of atoms and the BE matrix, which naturally describes covalent bonding and lone pairs. 
Each elementary step in a chemical reaction is represented by changes in the BE matrix (Fig.~\ref{fig1}b). The collective flow of electrons throughout the chemical transformation is formalized as the transformation of a probability distribution of electron localization from the reactants' state to the products' state. (Extended Fig.~\ref{fig1}a)
This design inherently prevents the creation or destruction of atoms and electrons during reaction prediction.

FlowER learns to analyze any state between reactants and products within an elementary step by featurizing the BE matrix and atom identities at a pseudo-timepoint $t$, where $t=0$ corresponds to the reactant state (prior) and $t=1$ corresponds to the product state (target). At its core is a graph transformer architecture \cite{vaswani2017attention, ying2021transformers} with an expressive multi-headed attention mechanism (Fig. \ref{fig1}c). Predicted electron movements, akin to partial arrow pushing, are applied to the reacting atoms to update the BE matrix for the next timepoint. 
Recursive predictions yield a full reaction mechanism step-by-step while ensuring that each intermediate state adheres to strict conservation principles. (Extended Fig.~\ref{fig1}b) 
The distributional nature of flow matching as a generative paradigm means that repeated sampling may produce distinct hypothetical products each time, allowing FlowER to propose branching mechanistic pathways, side products, and potential impurities. 

In order to train FlowER, we impute mechanistic pathways for a subset of the USPTO-Full dataset \cite{dai2019retrosynthesis} containing approximately 1.1 million experimentally-demonstrated reactions in patents from the United States Patent and Trademark Office. 1,220 expert-curated reaction templates were constructed for 252 well-described reaction classes, leading to a total of 1.4 million elementary reaction steps. Following the typical training procedure for conditional flow matching \cite{tong2023conditional}, interpolative trajectories sampled between reactant and product BE matrices are used as input, and the difference in the reactant-product BE matrices are used as the ground truth during model training. Additional details of training can be found in Methods.

Throughout this work, we compare FlowER to a prototypical sequence-based model, Graph2SMILES \cite{tu2022permutation}, which shares similar input features but proposes products directly as SMILES strings as a sequence-generation task. 
We evaluate Graph2SMILES as trained on traditional canonicalized SMILES strings (G2S) as well as on Kekulé structures with explicit hydrogens (G2S+H).

\subsection{FlowER resolves hallucinatory failure modes of sequence-based models by enforcing strict mass and electron conservation}\label{sec_conservation}

At the core of FlowER’s prediction is the $\Delta$BE matrix, which captures changes in electron configurations with a net sum of zero, thereby enforcing electron conservation. The $\Delta$BE matrix representation directly reflects the conventions of arrow-pushing diagrams, providing predictions that align with how chemists visualize reaction mechanisms. 
We try to capture the nuanced roles of electrons in chemical bonding and reactivity by further distinguishing lone pair and bond electron distributions. 
This design ensures that the prediction of the model is both physically consistent and interpretable, making a robust framework for understanding reactions at the mechanistic level.

\begin{figure*}[htbp]
\centering
\includegraphics[width=\textwidth]{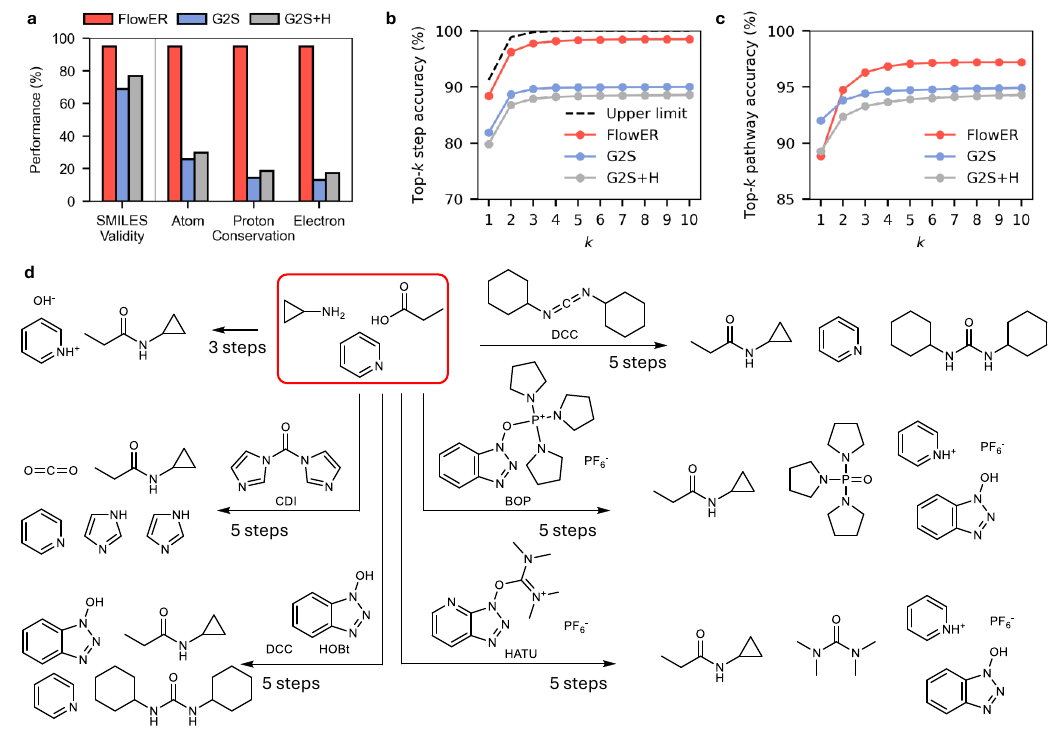}
\caption{\textbf{a}. Validity and conservation performance of FlowER compared to Graph2SMILES without (G2S) or with (G2S+H) Kekulé structures and explicit hydrogens. FlowER achieves higher rates of structure validity, heavy atom conservation, proton conservation, and electron conservation between reactants and products.
\textbf{b}. Top-\textit{k} step accuracy for predicting the product of a single elementary step. 
\textbf{c}. Top-\textit{k} pathway accuracy for predicting the full reaction pathway during beam search, showing the minimum beam width (\textit{k}) required to capture the correct sequence of elementary steps.
\textbf{d}. FlowER predictions and corresponding numbers of elementary steps for an example amide condensation under different reaction conditions: without a catalyst, using only DCC (N,N'-Dicyclohexylcarbodiimide), using DCC with HOBt (1-Hydroxybenzotriazole), using CDI (Carbonyldiimidazole), using BOP (Benzotriazol-1-yloxytris(dimethylamino)phosphonium hexafluorophosphate), and using HATU (O-(7-Azabenzotriazol-1-yl)-N,N,N',N'-tetramethyluronium hexafluorophosphate). FlowER successfully predicts the major products and byproducts for all six cases; some species are left activated and would be subsequently neutralized.}\label{fig2}
\end{figure*}

To demonstrate FlowER's unique ability to uphold the law of mass conservation, we evaluate models' predictions at the single elementary step level (Fig.~\ref{fig2}a).  
Valid intermediate or product SMILES strings are generated in approximately 95\% of test reactions. Valence rules are learned, not enforced, so 
structural anomalies like oxygens with four bonds are infrequently observed (details in Supplementary Section S3). 
The direct generation of SMILES tokens leads to substantially higher rates of invalid SMILES predictions (68.9\% valid), but could be partially mitigated through
Kekulé structures with explicit hydrogens (77.28\% valid) by virtue of their longer sequence length, providing more opportunities for the model to learn the syntax associated with generating valid SMILES.

While per-atom valence rules are not strictly enforced during generation, FlowER guarantees heavy atom, proton, and electron conservation as a direct consequence of its BE matrix representation. By contrast, despite being trained on a balanced mechanistic dataset (i.e., where the training data reflects conservation laws), sequence generative models violate conservation laws for the majority of predictions.
Specifically, only 31.4\% (G2S) and 30.1\% (G2S+H) of predictions maintain heavy atom conservation; when proton and electron conservation are also considered, the cumulative conservation rate drops to 14.3\% and 17.3\%.

These results highlight the critical failure of sequence-generation models to adhere to the fundamental principle of mass conservation even when trained on balanced datasets of more than one million elementary steps. Such inherent architectural limitations leads to nonsensical and hallucinatory failure modes and reduces confidence in these models when used by chemists. To underscore the premise of this work, FlowER inherently satisfies conservation laws, offering greater reliability, better user confidence, and predictions that are 
readily amenable to downstream tasks where precise mass and electron balances are essential, such as automated quantum chemical calculations as shown later.

\subsection{FlowER demonstrates high predictive accuracy on common reaction types and mechanistic pathways}\label{sec_accuracy}

Prediction of plausible reaction mechanisms is benefited by, but not guaranteed by, adherence to conservation principles.
We evaluate predictive performance in anticipating intermediates and final products across diverse chemical pathways  in terms of two metrics: top-\textit{k} step and top-$k$ pathway accuracies. To reiterate the distributional nature of conditional flow matching, FlowER can be sampled multiple times given the same reactants to yield multiple possible outcomes, which can then be ranked based on their frequency. Additional details of sampling can be found in Methods. 

The top-\textit{k} step accuracy measures the proportion of individual elementary steps  where the recorded product appears within the top-\textit{k} ranked predictions.
Multiple valid products can arise from the same reactants due to factors like variations in reaction conditions (e.g., different acids or bases) or in resonance forms, yet only one outcome is considered ``correct'' for each test reaction in our dataset. 
This results in an upper limit for top-\textit{k} step accuracy (Fig.~\ref{fig2}b, dashed black line). 
Beyond individual steps, the top-\textit{k} pathway accuracy evaluates the ability to identify complete reaction pathways through recursive application of elementary step prediction. Specifically, pathway accuracy measures the minimum beam width $k$ required for the model to reconstruct the correct sequence of intermediate steps leading to the final product. Pathway accuracy measures performance against experimentally-observed reaction outcomes from the patent literature and is not affected by any imperfections in our imputed mechanistic pathways.

For the 240 reaction types used in training, encompassing a total of 1,413,515 elementary steps across 251,796 reactions, FlowER demonstrates superior top-\textit{k} step accuracy compared to both G2S and G2S+H (Fig.~\ref{fig2}b). 
However, for top-1 pathway accuracy, FlowER achieves 88.8\% accuracy, slightly lower than G2S (92.0\%) and G2S+H (89.2\%) (Fig.~\ref{fig2}c).
In spite of this, FlowER surpassed both models in top-2 (and larger $k$) pathway accuracy, highlighting its substantially improved coverage when full, branching mechanistic pathways are generated (Supplementary Section S4).

Notably,  grounding product prediction in mechanistic understanding enables this high level of expressivity and performance with fewer parameters and fewer training examples. The performance comparisons in Fig.~\ref{fig2} are shown for a FlowER model with $7.2M$ parameters, compared to the $18.0M$ parameters required by G2S. 
Extended Data Fig.~\ref{train_size} demonstrates our ability to maintain strong predictive performance even with significantly reduced training data. When trained on only 500 elementary steps uniformly sampled from the original training set,
FlowER achieves top-1 and top-10 step accuracies of approximately 35\% and 40\%, respectively, compared to near-zero performance by G2S. 
As demonstrated later in Section~\ref{sec_fine_tuning}, this data efficiency extends to sample efficient fine-tuning on unseen reaction types with minimal additional training.

Beyond quantitative predictive accuracy and efficiency, qualitative analysis illustrates how FlowER captures the influence of specific reaction conditions into its mechanistic reasoning capabilities. Reaction conditions, such as temperature, concentrations, solvent, catalyst or reagent, and pH, play a critical role in determining the outcome of chemical reactions. These variables can influence reaction rates, selectivity, and overall yield. While FlowER does not explicitly account for continuous variables like temperature, concentration, or solvent polarity, it 
represents catalysts, solvents, and other agents in the BE matrix as participants in the reaction mechanisms.

As illustrated in Fig.~\ref{fig2}d, FlowER predicts the hypothetical condensation 
of a carboxylic acid and amine in the presence of pyridine as a base, accomplishing the reaction in three steps without the explicit definition of 
a coupling reagent. However, when coupling reagents such as DCC (N,N'-Ddicyclohexylcarbodiimide), DCC with HOBt (1-hydroxybenzotriazole), CDI (carbonyldiimidazole), HATU (O-(7-azabenzotriazol-1-yl)-N,N,N',N'-tetramethyluronium hexafluorophosphate), or BOP (benzotriazol-1-yloxytris(dimethylamino)phosphonium hexafluorophosphate) are introduced, FlowER predicts distinct five-step mechanisms leading to the same amide product but with different byproducts and activated intermediates. These predictions show FlowER’s ability to differentiate between reaction pathways based on the choice of reagents, reflecting subtle changes in mechanistic steps while preserving the consistency of the final major product.

\subsection{Predictions of elementary steps align with textbook chemistry principles describing reactivity and selectivity trends}\label{sec_acid_base}

Reactivity trends and heuristics, often derived from fundamental principles and corroborated by experimental results, play a key role in teaching and understanding reactivity. These principles enable chemists to reason bottom-up, starting from basic axioms to deduce likely reaction outcomes. 
Machine learning models instead rely on top-down statistical learning. Therefore, a natural question arises: do the same principles emerge from models trained only on reaction databases of experimental outcomes? Do such models 
reflect the underlying chemical reasoning encoded in human-developed heuristics? To address this, we probe FlowER’s predictions for elementary steps, examining whether they align with well-established textbook principles as a \emph{post hoc} analysis.

\begin{figure*}[htbp]
\centering
\includegraphics[width=\textwidth]{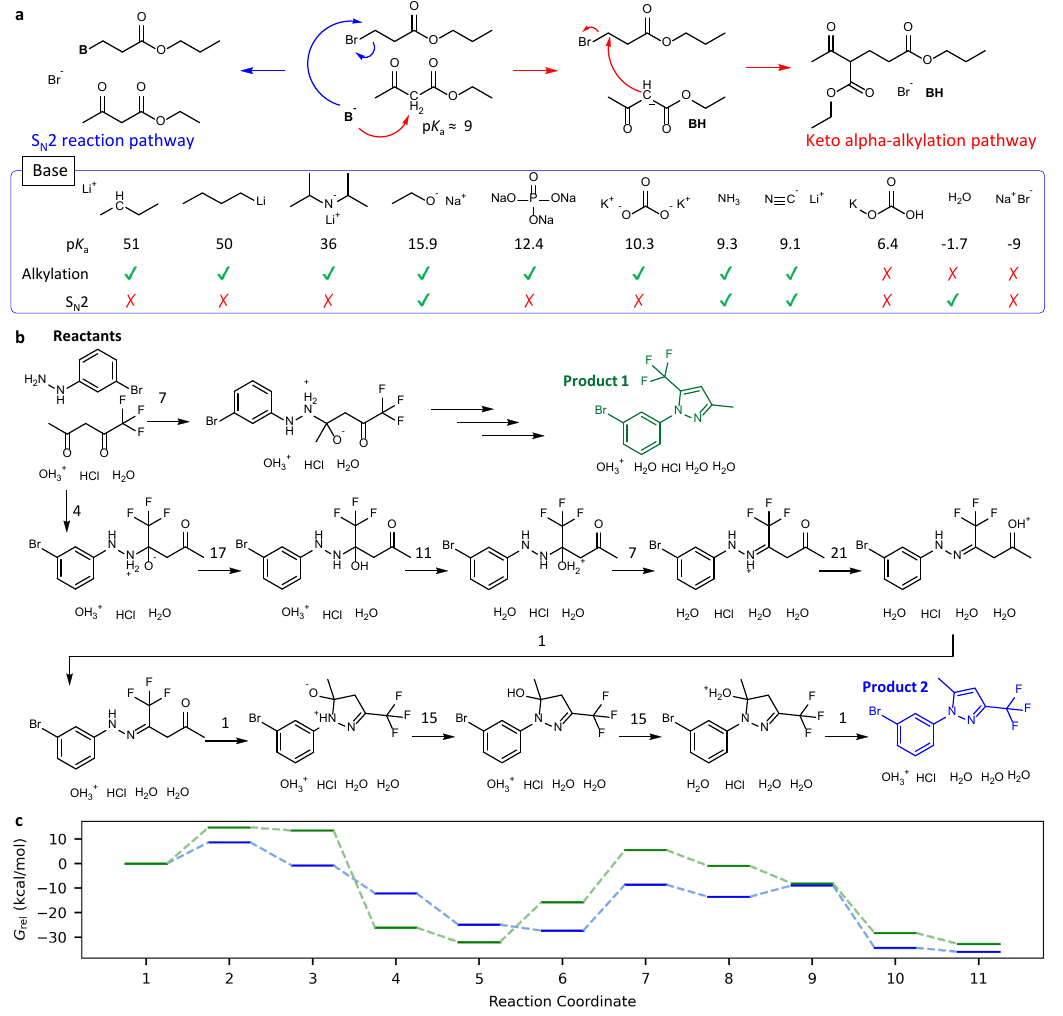}
\caption{
\textbf{a}. Analysis of FlowER predictions as a function of choice of base/nucleophile.
The reaction can proceed via two distinct pathways: the keto alpha-alkylation pathway and the S$_\text{N}$2 pathway. The choice between these pathways depends on whether the base functions primarily as a base or as a nucleophile. The blue box shows whether FlowER proposes the final product for each pathway when using various bases, annotated with their conjugate acid \pKa. If both pathways are marked with an `X', it means that FlowER predicted no reaction.
\textbf{b}. An example reaction reported in 
2024 \cite{US20240150295, US20240150296} (Extended Fig.~\ref{G2S_prediction}a), 
where FlowER successfully reproduces the two experimentally-recorded products in ten sequential steps. The numbers above each arrow represent the number of times that reaction was proposed during 32 independent sampling steps. The complete reaction pathway predicted by FlowER is provided in Extended Fig.~\ref{FlowER_prediction}.
\textbf{c}. \textit{G}$_{rel}$ values of each state, representing the Gibbs free energies, calculated using the B3LYP/6-311G level of theory with water as the solvent, modeled via the SMD method.The reaction coordinate spans from the initial reactant state at $x = 1$, to the final product state at $x = 11$, with intermediate steps corresponding to mechanistic transformations predicted by FlowER. The green pathway corresponds to the transformation leading to Product 1, while the blue pathway represents the route toward Product 2. All energy values are referenced relative to the reactant state set at 0 kcal/mol.}\label{fig3}
\end{figure*}

Many reactions are strongly influenced by---and reliant on---acid-base chemistry. 
At a textbook-level description, a reaction reliant on proton donation or abstraction may or may not proceed depending on the \pKa~values of the species involved. Acid-base reactions are prevalent in our dataset and were incorporated during mechanistic data curation, as mass conservation requires full bookkeeping of proton sources and sinks. 
This allows FlowER to express a data-driven understanding of basicity (Fig.~\ref{fig3}a). 

For example, the  $\alpha$-alkylation of $\beta$-ketoesters  
requires a base to deprotonate the $\alpha$-proton of the $\beta$-ketoester. The standardized \pKa~of this  $\alpha$-proton is set to 9 during data curation, regardless of substituent. However, if the base acts as a nucleophile, it can favor a competitive S\textsubscript{N}2 reaction over $\alpha$-alkylation. 
We examined FlowER's behavior across 16 unique bases. For the 11 bases shown in Fig.~\ref{fig3}a, those whose conjugate acid \pKa~exceeds 9 successfully led to the $\alpha$-alkylation product while weaker bases did not. 
Bases such as sodium ethoxide, ammonia, lithium cyanide, and water were perceived by FlowER to have sufficiently strong nucleophilicity to result in S\textsubscript{N}2 reaction products.

An expanded analysis of 29 bases with both neutral and ionic representations is summarized in 
Extended Table~\ref{acid_base_table}. 
Some bases exhibited differing reactivity between their neutral and ionic forms, while others showed consistent behavior across forms. For example, in the case of sodium ethoxide, the neutral form exclusively predicted keto alpha-alkylation, whereas the ionic form predicted both keto alpha-alkylation and S\textsubscript{N}2 pathways. 
For sodium ethanethiolate (EtSNa, \pKa~$\approx$ 10), neither the neutral nor ionic form successfully predicted $\alpha$-alkylation; instead both predicted the S\textsubscript{N}2 reaction. These nuances in behavior result from dataset trends (e.g., underrepresentation of thiols used as bases rather than nucleophiles) and representation conventions (e.g., of neutral sodium salts versus ionized fragments), both of which may be corrected through dataset expansion.

\subsection{FlowER predicts reaction pathways beyond training data through mechanistic reasoning}\label{sec_unrecognized}

One of our motivations for pursuing predictive chemistry models that operate at the mechanistic step level is to generalize to---and provide an explanation for---products that do not correspond to standard reaction types used for training. 
To explore this capability, we examined 22,000 reactions from patents reported in 2024 that were not assigned a specific reaction class in the Pistachio dataset \cite{pistachio}, potentially including multiple known transformations that occur in sequence or novel mechanistic pathways.
We performed a narrow beam search (width 2, depth 9) on 22,000 unrecognized reactions and successfully recovered 351 products. 
Several examples of FlowER-predicted pathways are provided in Supplementary Section S7.

One exemplary pathway is illustrated in Fig.~\ref{fig3}b. The experimental evidence for this reaction originates from two patents \cite{US20240150295, US20240150296}. This reaction was labeled as having an unknown reaction class due to the existence of two recorded products; although each corresponds to well-known pyrazole synthesis, FlowER did not see the Knorr pyrazole synthesis during training. Although reagents were not recorded in Pistachio, we added hydrochloric acid, water, and oxonium based on the acidic aqueous conditions described in the patents and expanded the beam search to  
width 5, depth 10.

FlowER's predicted 10-step mechanistic pathway begins with an initial nucleophilic attack, where 3-bromophenylhydrazine attacks an electrophilic $\beta$-diketone. Depending on which carbonyl group is attacked, the reaction can proceed via two distinct pathways. The nucleophilic attack forms a tetrahedral intermediate, followed by an intramolecular proton transfer. The next two steps involve generation of the required oxonium and a nitrogen-assisted condensation to form the corresponding iminium. Subsequent proton shuffling steps set the stage for a second nucleophilic attack and intramolecular cyclization.  
In the subsequent steps, two additional eliminations culminate in the final, experimentally-reported products.

In practical applications, this capability to generalize to new mechanistic sequences becomes especially valuable when experimental results deviate from expected outcomes. By analyzing the predicted stepwise pathway, chemists can pinpoint where deviations might have occurred, propose plausible explanations for unexpected products, and refine experimental conditions accordingly. Doing so with exact mass conservation also enables integration with downstream pipelines for automated computational feasibility assessment, for example, using 
density functional theory (DFT) calculations.  
To illustrate this use case, we present the free energies of intermediate states in Fig.~\ref{fig3}c, calculated at the B3LYP/6-311G level of theory with water as the solvent, modeled by the SMD method. 
This computational analysis on top of FlowER's predictions suggests that Product 2 (blue) is both thermodynamically and kinetically favored (assuming the Bell–Evans–Polanyi principle holds), aligning with the experimentally observed 8:2 ratio between Products 2 and 1 \cite{US20240150295, US20240150296}. 

The importance of mass conservation when incorporating quantum chemical calculations is further highlighted through beam search experiments with G2S under identical input conditions. As shown in Extended Fig.~\ref{G2S_prediction}b, although one of the pathways predicted by G2S reaches one of the major product (the other product is not recovered), its sequence violates mass conservation and precludes accurate thermodynamic calculations. This 
can be seen even in the top-2 ranked pathway, where the reaction terminates prematurely after E1-type elimination and also fails to conserve all atoms. Such inconsistencies are particularly prevalent when predicting outcomes of reactions dissimilar to training data.

\subsection{FlowER's understanding of mechanistic steps enables sample efficient fine-tuning on previously-unseen reaction types}\label{sec_fine_tuning}

\begin{figure*}[htbp]
\centering
\includegraphics[width=1\textwidth]{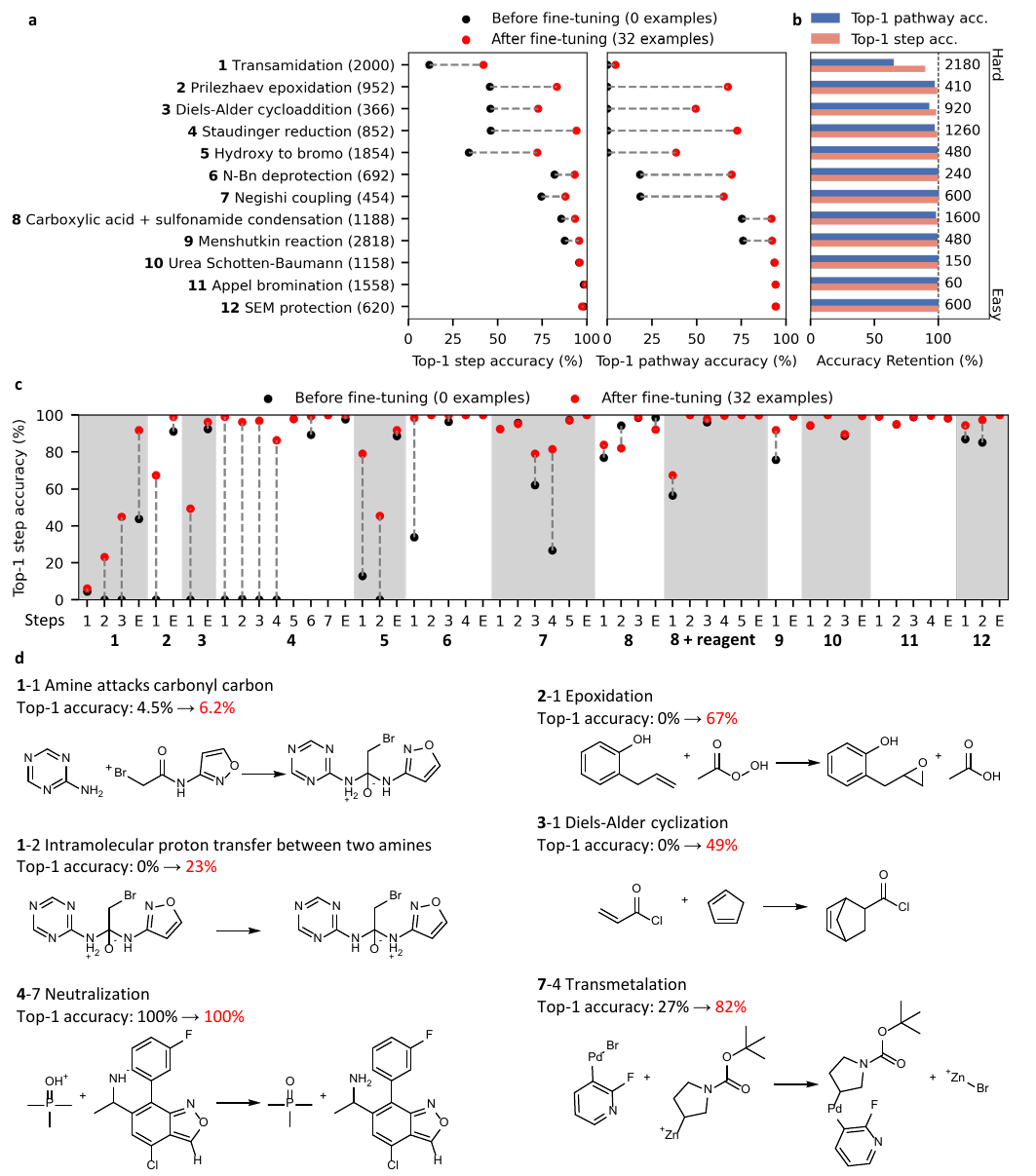}
\caption{\textbf{a}. Comparison of top-1 step accuracy and top-1 pathway accuracy before and after fine-tuning FlowER on reaction types absent from the training set. Fine-tuning was done using only 32 overall reactions per reaction type. Each unseen reaction type is annotated with the number of reactions used for evaluation. 
\textbf{b}. Accuracy of predictions on a 10\% random subset of the original test set after fine-tuning on unseen reaction types, normalized to performance of the base FlowER model. 
Each fine-tuned model is annotated with 
the number of steps used during fine-tuning with early stopping. 
\textbf{c}. Evaluation in top-1 step accuracy before and after fine-tuning for each reaction type’s elementary steps. Reaction \textbf{8} represents carboxylic acid + sulfonamide condensation, which is divided into two cases: one using coupling reagents and one without. ``E'' represents a pseudo-step where molecules remain unchanged at the end of a mechanistic sequence. 
\textbf{d}. Selected elementary steps and the their top-1 step accuracies before and after fine-tuning.}\label{fig4}
\end{figure*}

A fundamental premise of learning reaction mechanisms is the potential for better generalization to unseen reaction types that may share mechanistic steps with seen reaction types. That is, an ``extrapolation'' in terms of reaction type may truly be an ``interpolation'' in terms of mechanism. 
When the reaction proceeds through elementary steps that the model is not aware of, it must be retrained or fine-tuned. 
We evaluate FlowER on 12 out-of-distribution reaction types (Fig.~\ref{fig4}a). Performance varies significantly across these reaction types: for some, top-1 pathway accuracy exceeded 90\%, while for others, it was as low as 0\%. High-accuracy predictions tended to occur for reactions with elementary steps similar to those in the training set. 

Fine-tuning of reaction prediction models has traditionally made use of relatively large datasets, e.g., 
25,000 carbohydrate chemistry reactions \cite{pesciullesi2020transfer}, 9,959 Heck reactions \cite{wang2020heck}, or 2,254 Baeyer–Villiger reactions \cite{zhang2021data}. 
Here, we fine-tune FlowER on only 32 new reactions---orders of magnitude fewer than prior work---and achieve top-1 pathway accuracies above 80\% for 11 out of the 12 reactions. 
FlowER’s ability to adapt to such a minimal number of examples demonstrates its ability to draw analogy with known mechanistic steps and its robustness to scaffold variations within reaction types. 
Importantly, fine-tuning on specific reaction types previously not seen by the model does not lead to ``catastrophic forgetting'' \cite{luo2023empirical}
of the original breadth of reactions, as evidenced by preservation of 
accuracy on the original test set (Fig.~\ref{fig4}b). 
This result demonstrates FlowER's extensibility to additional reaction types while retaining its pretrained knowledge. 

We further analyze the top-1 step accuracy for each elementary step of each unseen reaction type (Fig.~\ref{fig4}c). For instance, type \textbf{5}, the Diels-Alder cycloaddition, consists of only two steps: the initial concerted cycloaddition and a second step where no further changes in the BE matrix occur, indicating reaction completion. 
Steps with high accuracy before fine-tuning are typically those well-covered in the training set by other reaction types. An example is step \textbf{4}-7 in Fig. \ref{fig4}d, the neutralization step in the Staudinger reduction, which exhibited 100\% accuracy even prior to fine-tuning. 
For other steps, it is understandable that they cannot be predicted well before fine-tuning. The transmetalation step in Negishi coupling (\textbf{7}-4) from zinc to palladium lacks any precedent in our USPTO-derived training dataset; the only examples of zinc participating in a reaction come from supplemental training examples from RMechDB and PMechDB \cite{tavakoli2023rmechdb, tavakoli2024pmechdb}, enabling 27\% step accuracy before fine-tuning.
Cycloaddition reactions (\textbf{3}-1) are completely novel prior to fine-tuning despite the model's familiarity with dienes and dienophiles in other reaction contexts. 

Transamidation proved to be the most challenging reaction type after fine-tuning. 
This three-step sequence requires (\textbf{1}-1) addition of an amine to the carbonyl carbon of an amide, (\textbf{1}-2) an intramolecular proton transfer between amines, and (\textbf{1}-3) elimination of the original amine. The model’s difficulty with this reaction may stem from biases learned during training. First, it has encountered numerous instances where amides and amines are present but do not react, leading it to associate these species with non-reactivity. Second, while the model has seen proton transfers from amine cations, they typically transfer to hydroxyls rather than to other amines. Finally, in cases with quaternary carbons, amines rarely act as a leaving group in the training data. These factors contribute to the model's challenge in learning and accurately predicting transamidation pathways. Nevertheless, for all other reaction types, FlowER successfully generalizes from its 32 fine-tuning examples to the hundreds or thousands of test examples in each test set.

\section{Conclusion}\label{Conclusion}

In this work, we establish a new formulation of the reaction prediction task that rigorously enforces mass and electron conservation. Revisiting classic BE matrix formalisms from Ugi, we leveraging the modern generative paradigm of flow matching to redefine reaction outcome prediction as a process of electron redistribution. In doing so, we provide a physically grounded and interpretable framework for modeling chemical reactivity. Our model, FlowER, captures mechanistic details at the elementary step level, resolving limitations in sequence-based models that often hallucinate atoms or violate conservation laws.

Empirical evaluations demonstrate that FlowER outperforms a prototypical sequence-generative approach in recovering standard mechanistic pathways. 
Additionally, the model’s robustness to data scarcity allows it to generalize effectively to previously unseen reaction types with minimal fine-tuning, illustrating its potential for data-efficient adaptation to novel chemistry domains. 
Beyond its quantitative predictive accuracy, FlowER exhibits alignment with textbook chemistry principles. We believe its mechanistic fidelity will make it a powerful tool for synthetic planning, reaction design, and computational recapitulation of experimental observations. Its conservation-aware formulation allows seamless integration with quantum chemical calculations, enabling thermodynamic and kinetic assessments of predicted pathways. As its training corpus of well-accepted mechanistic pathways expands, so too will the coverage of reaction types for which FlowER can make accurate predictions.

By strengthening the connection between machine learning-driven reaction modeling and fundamental mechanistic principles, we expect our modeling framework to create new opportunities for computational validation of mechanistic hypotheses and reaction discovery. Its ability to generate, corroborate, and explain reaction pathways opens new possibilities for improving data-driven reaction design and predictive chemistry applications.

\section{Methods}\label{Methods}
\subsection{Flow matching formulation for reaction outcome prediction}

We frame reaction outcome prediction as an electron redistribution process, where the initial reactant electron distribution transforms into the final product electron distribution. Following the discussion in Tong et al. \cite{tong2023conditional}, this transformation can be understood in the context of generative modeling, where the objective is to learn a mapping \( f: \mathbb{R}^d \to \mathbb{R}^d \) that pushes an initial distribution \( q_0 \) to a target distribution \( q_1 \). 
The mapping $f$ when extended to a time-dependent continuous transformation is known as an integration map \( \phi_t \) 
that defines a pushforward measure \( p_t := [\phi_t]_{\#} p_0 \), describing how the density of points from prior distribution are transported along the vector field $u_t(x)$ from \( t = 0 \) to \( t \). This measure is also commonly referred to as the probability path and satisfies the continuity equation,
\[
\frac{\partial p_t}{\partial t} = -\nabla \cdot (p_t u_t),
\]
ensuring the density's evolution over time is in accordance with the principle of mass conservation. In FlowER's formulation, time $t$ is bounded by $[0, 1]$, where $\rho_0$ is defined as a noised reactant electron distribution. We train a parametrized neural network model to learn the transformation towards $\rho_1$, the product electron distribution. Data points $x$ sampled from both reactant and product densities are represented using BE matrix. However, since the prior and target are both general distributions, the vector field $u_t(x)$ that generates the probability path $p_t(x)$ is unknown, thus intractable to compute. Thus, this motivates our use of  conditional flow matching (CFM) \cite{tong2023conditional}, which mitigates this issue of intractability. By conditioning on a variable $z$, from the data distribution $q(z)$, the desired unconditional probability path can be rewritten as an average conditional probability path with respect to the data distribution: $p_t(x) = \int p_t(x | z) p_1(z) \,dz$, paving the way to define a tractable loss for flow matching. In our formulation, $z=(x_0, x_1)$, a tuple of source and target samples corresponds to a pair of reactant-product BE matrices, together representing a complete reaction step.

The conditional probability path $p_t(x | z)$ can be described as the trajectory $\mu_t(z)$ of electron flow from reactant to product, which we defined as a linear interpolation with standard deviation  $\sigma_t$ moving from reactant to product BE matrix with respect to time.
In our case where the conditional probability path takes the form of a Gaussian, these transitional BE matrix sampled from the trajectory are perturbed using standard normal distribution. 

\[
   p_t(x|z) = \mathcal{N}(x | \mu_t(z), \sigma_t^2) =  
   \mathcal{N}(x | tx_1 + (1-t)x_0, \sigma_t^2)
\]

To guarantee the conservation of electrons on the BE matrix, where no electrons are added or removed during the reparameterization process and the matrix must remain symmetric, we introduce a symmetric and zero mean centered Gaussian distribution during sampling. This was inspired by EDM's \cite{hoogeboom2022equivariant} zero center of mass point clouds. 
Aside from enforcing electron conservation, symmetry ensures that the perturbation are consistent across same atom pairs on the BE matrix. This noising method is also applied when we sample $z \sim q(z)$, where we introduce stochasticity into $x_0$, the reactant BE matrix,  
to provide sampling diversity.

The conditional vector field $u_t(x|z)$ that generates the conditional probability path $\mu_t(x|z)$ has a unique closed-form expression when the latter takes the form of Gaussian. 
\[
u_t(x|x_0, x_1) = \frac{\sigma_t'}{\sigma_t}(x-\mu_t)+\mu_t'
                = x_1 - x_0
\]
This corresponds to the ``velocity'' of electrons moving across time in order to form the product, represented as $\Delta\text{BE matrix}$, a difference between product and reactant BE matrix (Fig.\ref{fig1}b). The core idea of flow matching is that the unconditional vector field $u_t(x)$ can be learned through an objective that leverages the conditional vector field $u_t(z)$,
\[
    \mathcal{L}_{CFM}(\theta) := \mathbb{E}_{t, q(z), p_t(x_t|z)}||v_\theta(t, x) - u_t(x|z)||^2
\]
where $t \sim U([0,1])$, $z \sim q(z)$ (i.e., a reaction sampled from the training distribution),  and $v_\theta(t, x)$ represents a parametrized neural network model (Fig.~\ref{fig1}c). The training objective is to regress against the time-dependent conditional vector field. At inference, we can
predict reaction outcomes of a noised reactant BE matrix $x_0$ by integrating the learned vector field from the parameterized model using off-the shelf numerical ODE solvers, i.e, an Euler method. Repeated sampling of the trained model can lead to a variety of products as a result of this addition of noise.

\subsection{Bond electron matrix design choices and featurization}
The bond electron (BE) matrix as formalized by Ugi describes the number of shared electrons between atoms and lone pairs of valence atoms.  
Since our objective is to simulate the reaction transformation while preserving the total sum of electrons, noise added during probability path density sampling is adjusted to be symmetric and sum to zero as mentioned above. The number of electrons in each entry of the BE matrix is expanded into a vector embedding using radial basis function (RBF) kernels  \cite{schutt2017schnet} centered at values evenly spaced between 0 and 8 in increments of 0.1  
to capture the granularity of BE matrix transition along the conditional probability path. 

The BE matrix used in FlowER diverges from Ugi’s original representation, which assigns a value of 1.5 to off-diagonal elements corresponding to aromatic bonds. Representing aromatic bonds as 3 electrons evenly shared between pairs of atoms can 
lead to electron overcounting in polycyclic compounds \cite{10.1007/BFb0051317, ugi1993computer}. 
Instead, FlowER employs a Kekulé representation, where aromatic bonds are converted to alternating single and double bonds. Details can be found in the SI. 

FlowER represents electrons as floating point numbers to better facilitate the training of CFM which operates on continuous values. 
This poses a problem when converting back from BE matrices to Lewis structure representations (or SMILES strings) of molecules, which relies on discrete values for bond representation. Naive rounding to integer values can alter the sum and therefore violate electron conservation. 
To bridge this gap, we employ sum-safe rounding \cite{calvin2018round} as a post-processing step to convert continuous BE matrix entries into discrete integers while maintaining the sum of electrons. 

After sum-preserving rounding, the bond electron (BE) matrix is transformed back into molecular graphs for evaluation. Additionally, we apply an optional validity enhancement based on established chemical rules, correcting any discrepancies in electron distribution based on the original valence electron counts on the reactant BE matrix. For example, carbon atoms are ensured to not exceed 4 bonds. This calibration increases the validity of the resulting SMILES by a few percentage points. 

\subsection{Mechanistic data preparation}
We use the USPTO-Full dataset as processed by Dai et al. \cite{dai2019retrosynthesis}, which contains 1,100,105 reactions as the foundation for our mechanistic dataset. Building on our previous methodology for imputing mechanistic pathways from experimentally-validated reactions \cite{joung2024reproducing}, we developed a new dataset encompassing 252 reaction classes and 185 distinct mechanisms.
Reaction types are labeled using  NextMove’s NameRxn software \cite{namerxn}. 

1,200 expert-curated reaction templates were constructed to describe the most common reaction mechanisms observed in the dataset as SMARTS strings. 
Reactions requiring protonation or deprotonation are encoded as unimolecular reaction templates. This avoids the need for individual SMARTS templates for each conjugate acid-base pair. Additionally, approximate \pKa~values are recorded for 88 distinct acid-base pairs. 
These templates were sequentially applied to the reactants of each reaction to impute intermediates and products, generating a detailed mechanistic pathway for each reaction. As in our previous work \cite{joung2024reproducing}, we pruned unproductive intermediates and retained only the mechanistic steps leading to the experimentally reported products.

Mechanistic data was manually and automatically reviewed and cleaned  
to enhance its consistency and reliability.  
Atom mappings were checked for uniqueness; BE matrices were verified to be symmetric and non-negative integers. 
We verified that the sum of the BE matrix values remains unchanged before and after each elementary reaction. 
This process resulted in the generation of 1,413,515 elementary steps for the training set, 15,744 for the validation set, and 159,573 for the test set. 

To augment this patent-derived dataset, we include Baldi's RMechDB  \cite{tavakoli2023rmechdb} and PMechDB \cite{tavakoli2024pmechdb}. These additional reactions were remapped with RXNMapper \cite{schwaller2021extraction}. 
We then applied the same cleaning protocols as used for our mechanistic dataset and merged the processed datasets. Ultimately, the final dataset consisted of 250,782, 2,801, and 28,049 overall reactions, corresponding to 1,445,189, 15,744, and 162,002 elementary steps for the training, validation, and testing of FlowER.

\backmatter

\bmhead{Supplementary information}

Supplementary information is available for this manuscript, which includes Supplementary Sections S1-S8, Supplementary Figs. S1-S23, Supplementary Table S1, and additional references.

\section*{Declarations}

\subsection*{Acknowledgements}
We thank our colleagues in the Machine Learning for Pharmaceutical
Discovery and Synthesis consortium for help discussions regarding this work. 
The authors acknowledge the MIT SuperCloud and Lincoln Laboratory Supercomputing Center that have contributed to the research results reported within this paper.

\subsection*{Funding }
This work was supported by the Machine Learning for Pharmaceutical Discovery and Synthesis consortium and the National Science Foundation under Grant No. CHE-2144153. 

\subsection*{Competing interests}

The authors declare no competing interests.

\subsection*{Data availability }

The data used in this manuscript and all of the pretrained single-step models are available on Figshare at \href{https://doi.org/10.6084/m9.figshare.28359407.v2}{https://doi.org/10.6084/m9.figshare.28359407.v2}.

The full train, validation, and test sets for the 12 out-of-distribution reaction types used in \ref{sec_fine_tuning} are available upon request due to license considerations of NameRxn from NextMove Software.

\subsection*{Materials availability}

Not applicable

\subsection*{Code availability }

Source code for curating the dataset can be found in the following GitHub repository: \href{https://github.com/jfjoung/Mechanistic_dataset}{https://github.com/jfjoung/Mechanistic\_dataset}. Source code for FlowER can be found in the following GitHub repository: \href{https://github.com/FongMunHong/FlowER}{https://github.com/FongMunHong/FlowER}. 
These repositories include a detailed README with instructions on how to reproduce the results in this manuscript. 

\subsection*{Author contribution}

J.F.J., M.H.F., N.C., and C.W.C. conceptualized the project. J.F.J. and M.H.F. developed the methods and conducted computational experiments. J.F.J., M.H.F., and N.C. analyzed the results and prepared the original draft. J.F.J., J.P.L., and N.S.D curated the data. All authors reviewed and approved the manuscript.

\noindent






\setcounter{figure}{0}
\renewcommand{\figurename}{Extended Data Fig.}
\renewcommand{\theHfigure}{Extended Data Fig.\arabic{figure}}
\renewcommand{\tablename}{Extended Data Table} 
\renewcommand{\theHtable}{Extended Data Table\arabic{figure}} 
\renewcommand{\vfill}{\vspace*{\fill}}

\newpage
\begin{appendices}

\begin{table*}[h!]
\centering
\caption{FlowER's reaction predictions based on the \pKa~of the conjugate acid of various bases as shown in Figure~\ref{fig3}a. A total of 16 products were sampled for each set of reactants. ``Alkylation'' refers to the keto alpha-alkylation pathway, where the numbers separated by vertical bars represent the number of successful predictions for individual elementary steps: deprotonation, alkylation, and the end of the reaction. ``S\textsubscript{N}2'' indicates predictions for the S\textsubscript{N}2 reaction and the end of the reaction. A dash (-) represents that the corresponding pathway was not predicted.}
\label{acid_base_table}
\resizebox{\linewidth}{!}{
\begin{tabular}{|l|c|c|c|l|c|c|c|}

\bottomrule
\textbf{Base} & \textbf{\pKa} & \textbf{Alkylation} & \textbf{S$_\text{N}$2} & \textbf{Base} & \textbf{\pKa} & \textbf{Alkylation} & \textbf{S$_\text{N}$2} \\ 
\hline \hline
$n$-Bu$^-$ Li$^+$ & 50 & 4|12|15 & - & HCO$_3^-$ K$^+$ & 6.4 & - & 2|13 \\ 
$n$-BuLi & 50 & 10|9|14 & - & HCO$_3$K & 6.4 & - & - \\ 
$s$-Bu$^-$ Li$^+$ & 51 & 1|11|15 & - & 2K$^+$ CO$_3^{2-}$ & 10.3 & 6|14|16 & - \\ 
$s$-BuLi & 51 & 8|8|16 & - & K$_2$CO$_3$ & 10.3 & - & - \\ 
LDA & 36 & 7|15|16 & - & Na$^+$ Br$^-$ & -9 & - & - \\ 
Ionic LDA & 36 & 13|14|16 & - & NaBr & -9 & - & - \\ 
EtO$^-$ Na$^+$ & 15.9 & 2|14|16 & 8|16 & Na$^+$ OH$^-$ & 15.9 & 11|14|15 & 1|16 \\ 
EtONa & 15.9 & 7|15|16 & - & NaOH & 15.9 & 2|15|16 & 2|16 \\ 
NC$^-$ Li$^+$ & 9.1 & 2|9|16 & 1|16 & 3Na$^+$ PO$_4^{3-}$ & 12.4 & 7|15|16 & 1|15 \\ 
NCLi & 9.1 & 6|9|15 & - & Na$_3$PO$_4$ & 12.4 & 2|14|16 & - \\ 
EtS$^-$ Na$^+$ & 10 & - & 14|16 & 2Na$^+$ HPO$_4^{2-}$ & 7.2 & - & 4|16 \\ 
EtSNa & 10 & - & 2|16 & Na$_2$HPO$_4$ & 7.2 & - & - \\ 
NH$_3$ & 9.3 & 4|9|15 & 1|16 & Na$^+$ H$_2$PO$_4^-$ & 2.1 & - & 1|15 \\ 
H$_2$O & -1.7 & - & 1|16 & NaH$_2$PO$_4$ & 2.1 & - & - \\ 
DBU & 13.5 & 11|16|16 & 3|16 & & & & \\ 
\bottomrule

\end{tabular}}
\end{table*}

\newpage
\begin{figure*}[htbp]
\centering
\includegraphics[width=1\textwidth]{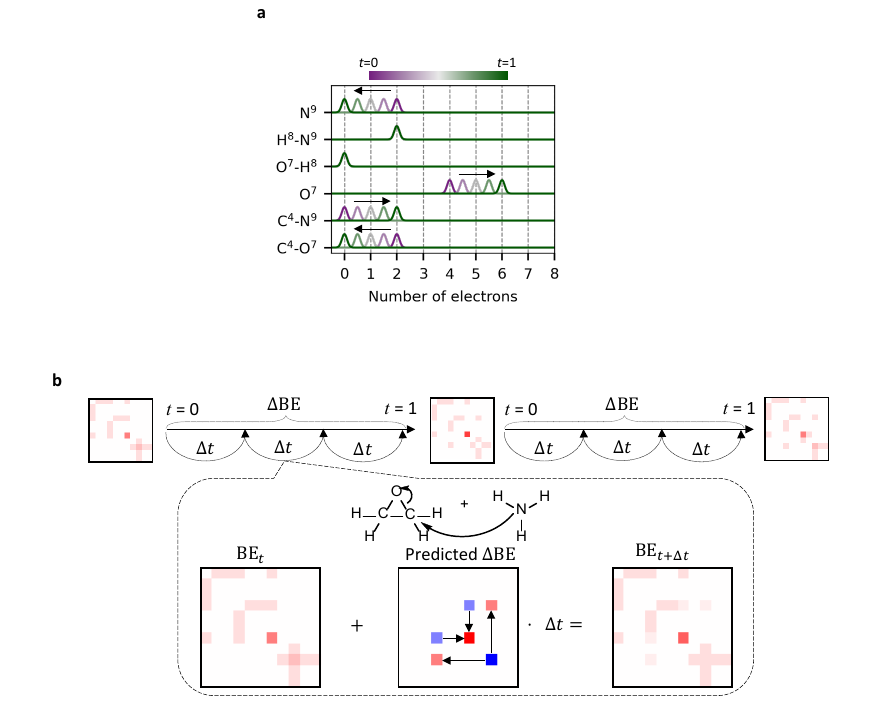}
\caption{\textbf{a}. Flow matching is a recently-developed generative technique \cite{tong2023conditional} that is well suited to learning the process of electron redistribution. This diagram illustrates the flow matching training process for the reaction of epoxide and amine and it's electron distribution in Fig.\ref{fig1}b. It highlights the changes of electron count on lone pairs and bonds, steered by a vector field (black arrow), from $t=0$ (reactant) to $t=1$ (product). 
\textbf{b}. Provided a set of reacting species represented by their bond-electron matrix, a trained model predicts the redistribution of electrons in an iterative manner that ultimately leads to the prediction of one mechanistic step. This can be repeated to provide a full multi-step sequence resulting in a stable product.
} \label{inference}
\end{figure*}

\newpage
\begin{figure*}[htbp]
\centering
\includegraphics[width=1\textwidth]{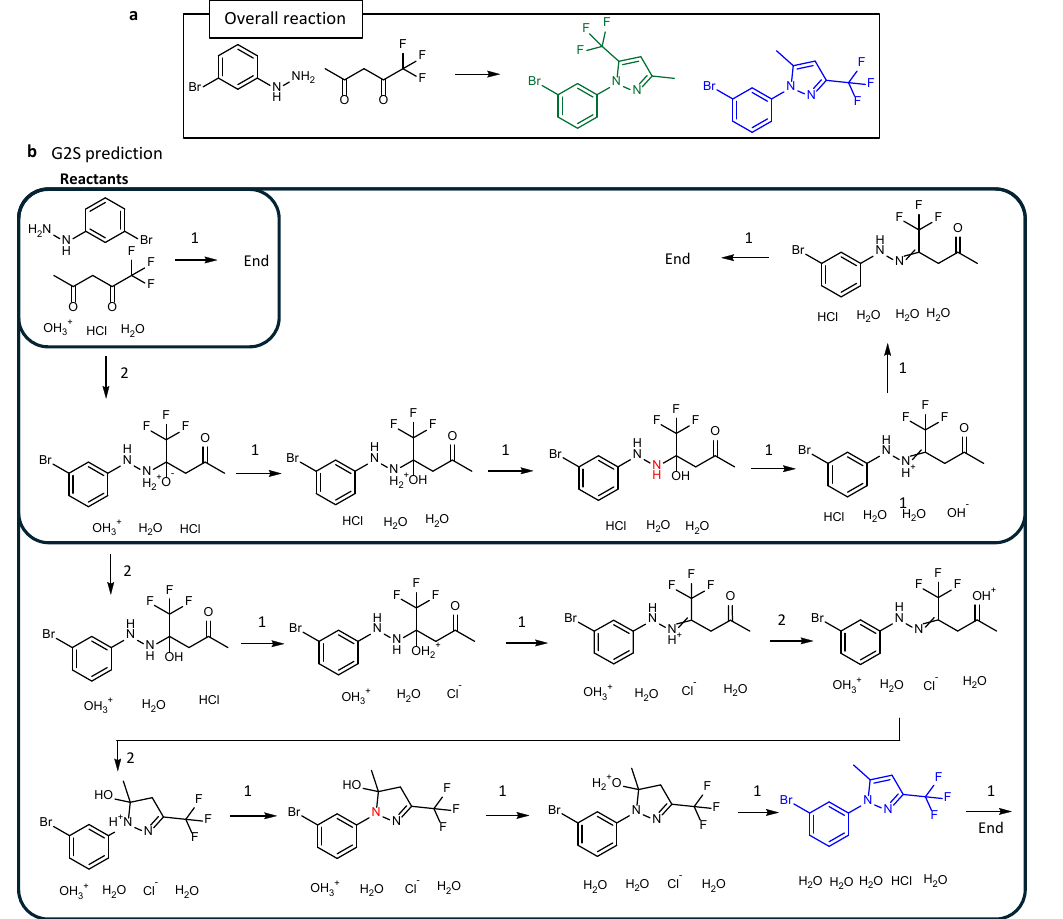}
\caption{\textbf{a}. An experimental reaction example reported in the patent literature in 2024 \cite{US20240150295, US20240150296}, which corresponds to a reaction type not 
seen during training.
\textbf{b}. G2S predicts top 1 and 2 mechanistic sequence that either remains unreactive or terminates prematurely. Although G2S manages to predict one of the major product through a lower ranked pathway, it still exhibits ``alchemy'' which violates mass conservation (highlighted in red where violations occur). The numbers above each arrow represent the rank of that reaction as proposed by G2S.}\label{G2S_prediction}
\end{figure*}

\newpage
\begin{figure*}[htbp]
\centering
\includegraphics[width=1\textwidth]{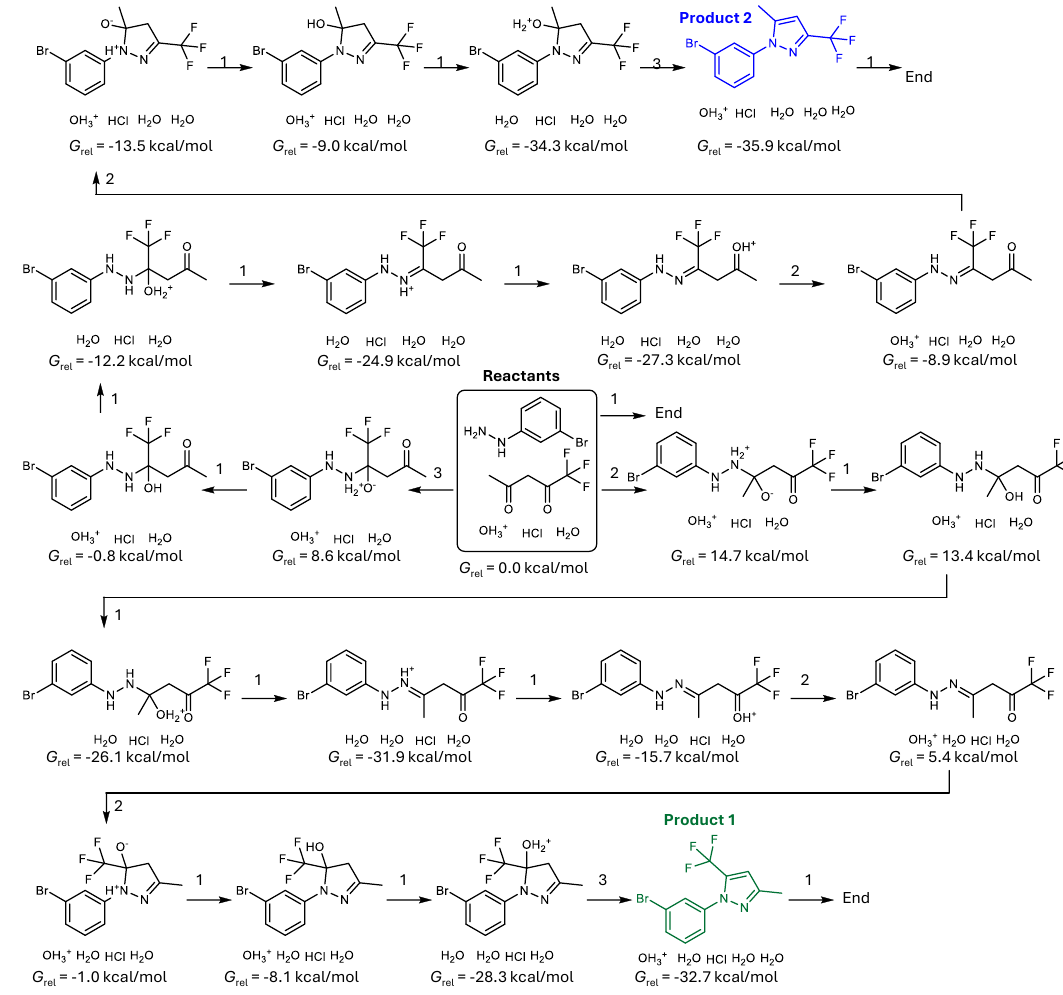}
\caption{ An example reaction reported in 
2024 \cite{US20240150295, US20240150296}, 
where FlowER successfully reproduces the two experimentally-recorded products in ten sequential steps. The numbers above each arrow represent the rank of that reaction as proposed by FlowER. The top 1 mechanistic sequence pathway remains unreactive, similar to G2S's prediction. }\label{FlowER_prediction}
\end{figure*}

\newpage
\begin{figure*}[htbp]
\centering
\includegraphics[width=1\textwidth]{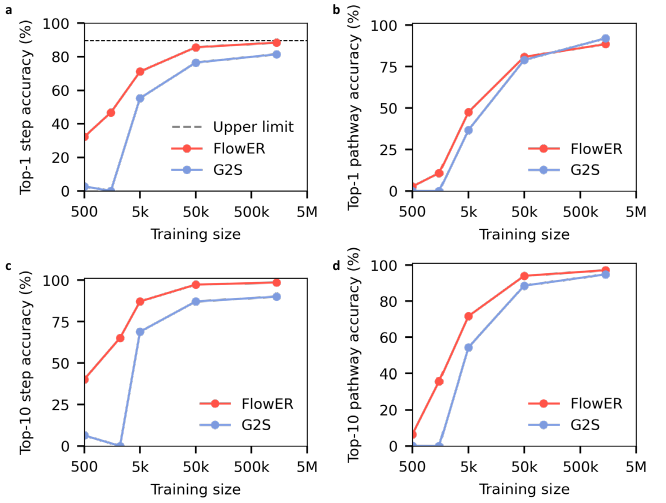}
\caption{In-distribution performance as a function of training size. Subsets of the full USPTO-Full derived training set are sampled at random for each training size; FlowER and G2S are trained and evaluated on the same splits. \textbf{a}. Top-1 step accuracy. \textbf{b}. Top-1 pathway accuracy. \textbf{c}. Top-10 step accuracy.  \textbf{d}. Top-10 pathway accuracy. }\label{train_size}
\end{figure*}

\newpage



\end{appendices}


\bibliography{sn-bibliography}

\clearpage
\includepdf[pages=-]{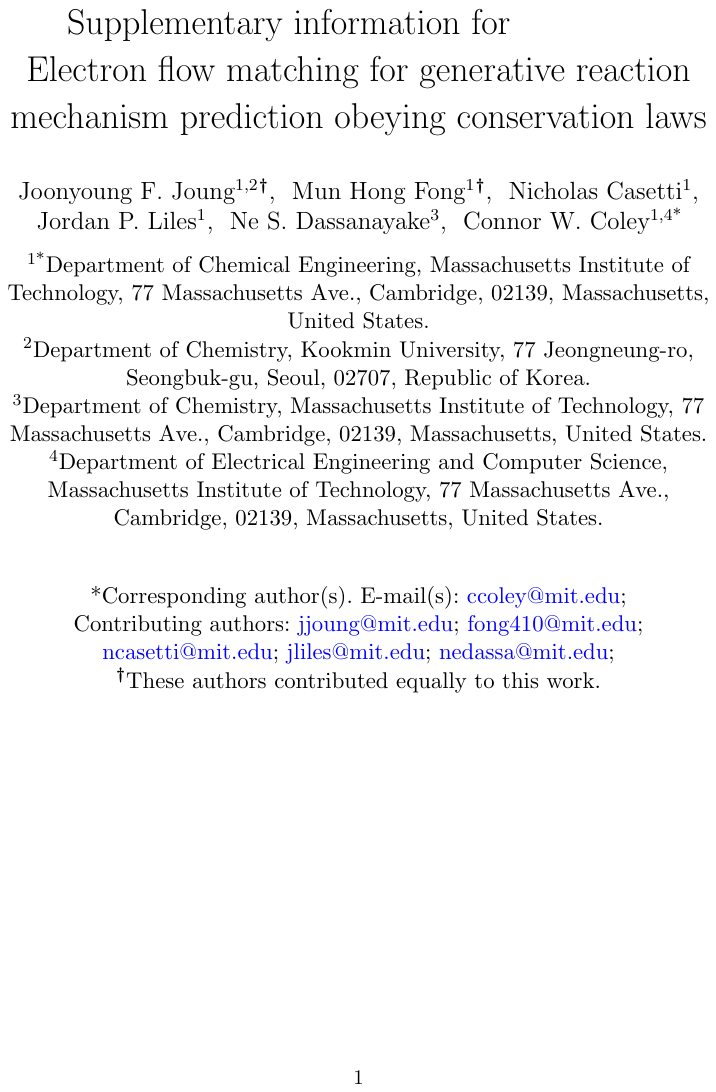}

\end{document}


\title[Article Title]{Supplementary information for\\ Electron flow matching for generative reaction mechanism prediction obeying conservation laws}

\author[1,2]{\fnm{Joonyoung F.} \sur{Joung}}\email{jjoung@mit.edu}
\equalcont{These authors contributed equally to this work.}
\author[1]{\fnm{Mun Hong} \sur{Fong}}\email{fong410@mit.edu}
\equalcont{These authors contributed equally to this work.}

\author[1]{\fnm{Nicholas} 
\sur{Casetti}}\email{ncasetti@mit.edu}

\author[1]{\fnm{Jordan P.} \sur{Liles}}\email{jliles@mit.edu}
\author[3]{\fnm{Ne S.} \sur{Dassanayake}}\email{nedassa@mit.edu}

\author*[1,4]{\fnm{Connor W.} \sur{Coley}}\email{ccoley@mit.edu}

\affil*[1]{\orgdiv{Department of Chemical Engineering}, \orgname{Massachusetts Institute of Technology}, \orgaddress{\street{77 Massachusetts Ave.}, \city{Cambridge}, \postcode{02139}, \state{Massachusetts}, \country{United States}}}

\affil[2]{\orgdiv{Department of Chemistry}, \orgname{Kookmin University}, \orgaddress{\street{77 Jeongneung-ro, Seongbuk-gu}, \city{Seoul}, \postcode{02707}, \country{Republic of Korea}}}

\affil[3]{\orgdiv{Department of Chemistry}, \orgname{Massachusetts Institute of Technology}, \orgaddress{\street{77 Massachusetts Ave.}, \city{Cambridge}, \postcode{02139}, \state{Massachusetts}, \country{United States}}}

\affil[4]{\orgdiv{Department of Electrical Engineering and Computer Science}, \orgname{Massachusetts Institute of Technology}, \orgaddress{\street{77 Massachusetts Ave.}, \city{Cambridge}, \postcode{02139}, \state{Massachusetts}, \country{United States}}}

\maketitle

\clearpage
  
\tableofcontents

\setcounter{figure}{0}
\renewcommand{\figurename}{Fig.}
\renewcommand{\theHfigure}{Fig. S\arabic{figure}}
\renewcommand{\tablename}{Table} 
\renewcommand{\theHtable}{Table S\arabic{figure}} 
\renewcommand{\vfill}{\vspace*{\fill}}

\newpage 
\section{Code and Data availability}
All data used in this study, along with pretrained single-step models, are publicly available on Figshare:  
\href{https://doi.org/10.6084/m9.figshare.28359407.v2}{https://doi.org/10.6084/m9.figshare.28359407.v2}.  
The dataset is provided in a text format, including reaction SMILES.

The source code for dataset curation can be found in the following GitHub repository:  
\href{https://github.com/jfjoung/Mechanistic_dataset}{https://github.com/jfjoung/Mechanistic\_dataset}.  
This repository includes scripts for preparing mechanistic dataset.

The implementation of FlowER, including model training, inference, and evaluation, is available at:  
\href{https://github.com/FongMunHong/FlowER}{https://github.com/FongMunHong/FlowER}.  
This repository provides training scripts and inference pipelines necessary to reproduce the results presented in this manuscript.

Both repositories include a detailed README with instructions for installation, usage, and result reproduction.

\newpage 
\section{Mechanistic dataset preparation}\label{si_dataset}
We utilized the USPTO-Full dataset \cite{dai2019retrosynthesis}, which contains 1,100,105 reactions, for curating an expanded mechanistic dataset. Building on our previous method for constructing mechanistic datasets \cite{joung2024reproducing}, we significantly extended the scope of the dataset. This involved creating 1,220 unique reaction templates, covering 252 reaction classes and 185 distinct mechanisms. By adding more reaction classes and mechanisms, we ensured broader coverage and diversity. Since our method of data imputation relies on reaction class labels, we classified reactions using NextMove’s NameRxn software \cite{namerxn}.

\subsection{Expert-curated reaction templates}
Expert-curated reaction templates were constructed to represent the most common reaction mechanisms observed in the USPTO-Full dataset. Each template was designed to encode the specific chemical transformations of an elementary step. To promote the accuracy of the constructed templates, three chemists reviewed each mechanism. A series of templates corresponding to a given reaction class was then sequentially applied to a set of reactants, generating possible intermediates and ultimately the products.

To ensure consistency with the experimentally observed products of the overall reactions, we algorithmically pruned unproductive intermediates and retained only the mechanistic steps leading to the reported products. Through this process, we were able to impute intermediates not explicitly recorded in the overall reactions and identify often-overlooked byproducts.

\subsection{Stoichiometry and acid-base reactions}

A significant number of elementary steps involve acid-base reactions. In our previous work \cite{joung2024reproducing}, hydrogen conservation was not fully addressed due to the challenge of explicitly specifying all possible acid-base partners for each template. To overcome this limitation, we redesigned the templates to ensure hydrogen conservation and accurately model acid-base reactions. Each template was constructed as a unimolecular reaction, where the intermediate either gains or loses a proton. To account for the proton donor or acceptor, we annotated the \pKa of the required partner directly in the template.

We compiled \pKa values for 88 distinct acid-base pairs. During the mechanistic dataset generation process, whenever a template requiring a proton transfer was applied, the following procedure was implemented:

\begin{itemize}
    \item If the reaction required an acid, the system checked whether an acid with a \pKa lower than the specified value was present in the current set of chemicals. If such an acid was found, it was added to the reactant side of the reaction, while the corresponding base was added to the product side, transforming the unimolecular template into a bimolecular one.
    \item Conversely, if a base was required, the system checked for a base with a \pKa higher than the specified value and adjusted the reaction in the same manner.
\end{itemize}

This dynamic adjustment allowed us to avoid creating exhaustive templates for every possible acid-base reaction, significantly streamlining the dataset construction process.

Additionally, while applying reaction templates, we addressed scenarios where multiple equivalents of a reactant were required, a common occurrence in acid-base and catalytic reactions. For example, the USPTO-Full dataset often records only one equivalent of a reactant, even when two or more equivalents are necessary for the reaction to proceed. If a template required the same reactant that had already been consumed in a previous elementary step, the mechanistic data generation process was adjusted as follows:

\begin{enumerate} 
    \item All earlier elementary steps were revised to duplicate the required reactant, ensuring sufficient quantities were present throughout the reaction sequence. 
    \item The current template was then applied to the reaction, utilizing the additional equivalent of the reactant. 
\end{enumerate}

This approach ensured that the mechanistic pathways generated from the templates remained chemically accurate, consistent with stoichiometric requirements, and reflecting the realistic reaction conditions.

\subsection{Kekulé representation of molecules}

\begin{figure*}[h]
\centering
\includegraphics[width=\textwidth]{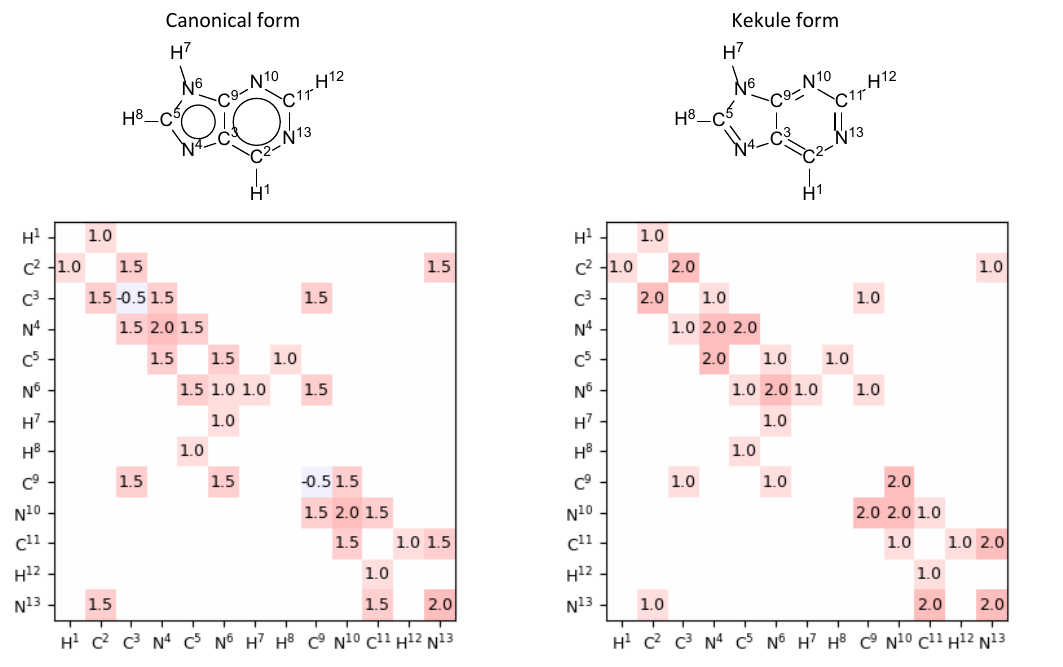}
\caption{
Comparison of BE matrices for the canonical and Kekulé forms. The canonical form represents aromatic bonds with bond order 1.5, leading to non-physical fractional electron counts. In contrast, the Kekulé form explicitly represents alternating single and double bonds, ensuring integer number of electrons.
}\label{fig_kekule}
\end{figure*}

In constructing the BE matrix, we adopted the Kekulé representation for aromatic rings instead of using the classical BE matrix form proposed by Ugi in the 1970s \cite{10.1007/BFb0051317, ugi1993computer}. In Ugi’s BE matrix, aromatic bonds are represented with a bond order of 1.5. However, in fused ring systems, this approach leads to an issue where certain carbon atoms, shared between two aromatic rings, end up with three bonds of order 1.5. As shown in the BE matrix of the canonical form in Fig. \ref{fig_kekule}, this results in a total electron count of 4.5 for such atoms, requiring lone pairs to be assigned a non-physical value of $-0.5$ electrons to maintain overall electron conservation. While the total electron count remains consistent, the introduction of negative fractional electrons is unrealistic and undesirable.

To address this, we adopted the Kekulé form, explicitly representing aromatic bonds as alternating single and double bonds. As seen in the BE matrix of the Kekulé form in Fig.~\ref{fig_kekule}, this approach ensures that all bond orders are represented as non-negative integers, making the BE matrix more physically meaningful. Kekulization was applied systematically using RDKit’s \texttt{Chem.Kekulize(mol)} function.

\subsection{Data cleaning and validation}
To ensure the reliability and consistency of the mechanistic dataset, we applied rigorous cleaning procedures.

\begin{itemize}
    \item \textbf{Atom Mapping Consistency}: All atom mappings were verified to be unique and consistent between reactants and products. 
    \item \textbf{BE Matrix Validation}: The BE matrix was checked to be diagonally symmetric, and all entries were ensured to be either 0 or positive integers. Conservation of electrons was verified by confirming that the sum of the BE matrix values remained constant before and after each reaction.
    \item \textbf{Chemical Validity Check}: Structures were screened using RDKit to remove chemically invalid molecules, such as those with incorrect valency or problematic resonance structures.
    \item \textbf{Kekulization Filtering}: If Kekulization failed for a given molecule, all steps involving that molecule were removed from the dataset.
    \item \textbf{Pathway Integrity Check}: If the removal of invalid steps resulted in the loss of a complete reaction pathway from reactants to the observed product, the entire sequence was discarded to maintain dataset consistency.
    
\end{itemize}

These validation steps ensured that the final dataset contained only chemically meaningful and computationally robust reaction sequences, improving its reliability for reaction modeling.

\subsection{Integration with external databases}
To enhance dataset diversity, we incorporated data from RMechDB \cite{tavakoli2023rmechdb} and PMechDB \cite{tavakoli2024pmechdb}. Since these external datasets provided atom mappings only for reaction centers, we used RXNMapper \cite{schwaller2021extraction} to extend atom mappings to the entire reaction while ensuring consistency with our existing mapping approach.

Before integration, we applied the same cleaning protocols used for our mechanistic dataset, including atom mapping validation, BE matrix verification, and stoichiometry checks.

\subsection{Final dataset}
The resulting dataset consisted of:

\begin{itemize}
\item Training Set: 250,782 overall reactions (1,445,189 elementary steps)
\item Validation Set: 2,801 overall reactions (15,744 elementary steps)
\item Test Set: 28,049 overall reactions (162,002 elementary steps)
\end{itemize}

The dataset spans 252 reaction classes and 185 distinct mechanisms, covering a broad range of organic transformations. The dataset was split into training, validation, and test sets using an 89:1:10 ratio.

\clearpage
\section{Post processing of model output}\label{si_post_processing}
\subsection{Sum-safe rounding}
The sum-safe rounding is performed through the python package \texttt{iteround} \cite{calvin2018round}. It provides a safe-sum rounding, which ensures the preservation of electron counts after rounding from floating point number back to representable integer bond order on the BE matrix, allowing us to reconstruct molecules back using RDKit.
\[
    \text{diff} = x - round(x)
\]
This equation represents how each value is incremented / decreased sequentially according to the sorted difference. The algorithm alternates between the smallest surplus / deficit to alter each floating point back to it's rounded integer.

\subsection{Validity Fix}
The validity fix stems from the notion that the product atoms' valence electron count should be the same as the reactant atoms' valence electron count. E.g., if the input reactant valence electron count is [2, 8, 8, 8, 8, 2], predicted product valence electron count is [2, 6, 8, 8, 10, 2], difference would be [0, -2, 0, 0, +2, 0]. If the sum of difference is zero, we apply this validity fix through ``redistributing" the electrons back to match it's reactant valence electron count. Empirically, this fix improves BE matrix to SMILES conversion validity by 3-4\% on the test set, and typically reactants with HCNOF elements that strictly obey the octet rule benefits from it.

\subsection{Failure mode analysis}
\begin{figure}[h]
\centering
\includegraphics{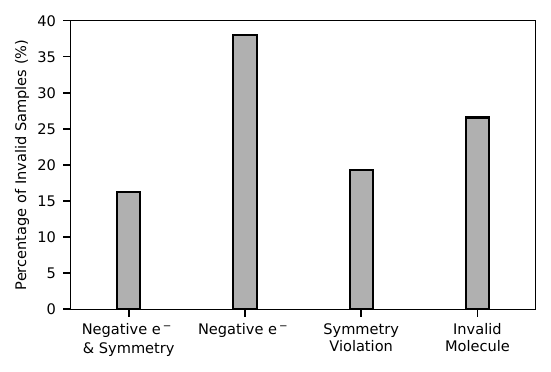}
\caption{Histogram of failure modes in FlowER's sampled predictions. The four categories represent different causes of invalid SMILES: (1) negative electron counts and symmetry violation, (2) negative electron counts only, (3) symmetry violation only, and (4) chemically invalid molecules. 
}\label{failure}
\end{figure}

Despite the post-processing techniques described above, FlowER still produces 5.06\% invalid SMILES during sampling, as mentioned in the main text. To better understand these failure cases, we analyzed the invalid outputs and categorized them into three major types. 
Fig.~\ref{failure} illustrates the distribution of the three failure cases identified in FlowER’s sampled predictions.

\begin{figure*}[h]
\centering
\includegraphics[width=\textwidth]{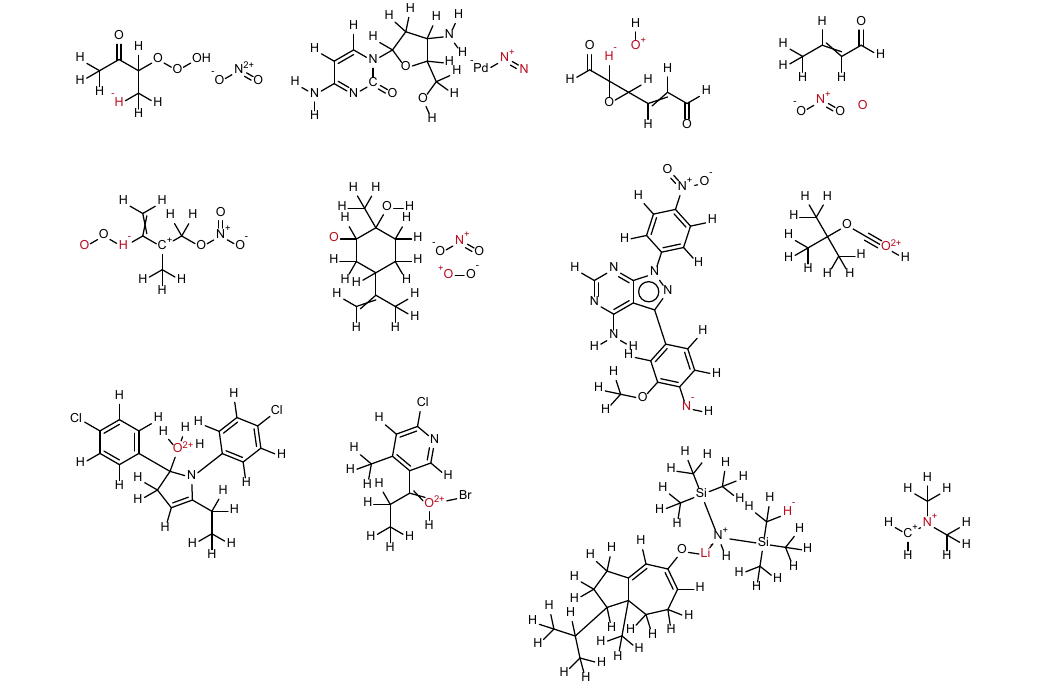}
\caption{Examples of chemically invalid molecules generated by FlowER. Problematic regions within the molecules are highlighted in red. 
}\label{weird_molecule}
\end{figure*}

\begin{itemize}
    \item \textbf{Electron over-redistribution in prediction}

    In some cases, FlowER predicts an excessive electron redistribution, moving more electrons from the bonds/lone pairs than what is originally  available in the reactants. This results in negative electron counts in the product BE matrix, making it physically invalid. This issue originates from errors in the $\Delta$BE matrix prediction, where the model incorrectly estimates the number of electrons involved in rearrangement.

    \item \textbf{Diagonal symmetry violation after sum-safe rounding}

    While sum-safe rounding ensures that the total number of electrons is preserved, it does not guarantee that the product BE matrix remains diagonally symmetric. If the rounding process distorts symmetry, the number of bonding electrons can become odd-valued, which leads to errors when reconstructing molecules. 
    
    In BE matrix-based molecule reconstruction, single, double, and triple bonds are defined as having 2, 4, and 6 bonding electrons, respectively. If a bond has an odd number of electrons due to symmetry breaking, it cannot be correctly mapped back to a valid molecular structure, resulting in an invalid SMILES.

    \item \textbf{Chemically invalid molecule despite valid BE matrix}

    In some cases, the sampled product BE matrix maintains mass conservation and diagonal symmetry but still fails to generate a chemically valid molecule. These cases occur when the predicted electron distribution leads to an unstable molecular structure, violating chemical stability rules. Examples of such chemically invalid molecules are illustrated in Fig.~\ref{weird_molecule}.
    
\end{itemize}

\clearpage
\section{Elaboration of pathway accuracy comparison}
Accurate prediction of chemical reaction pathways requires not only identifying correct individual mechanistic steps but also reconstructing the full sequence leading to the final product. Pathway accuracy measures a model’s ability to correctly predict complete reaction sequences, rather than individual steps. This is particularly important for FlowER, where each elementary step prediction influences subsequent steps, making pathway reconstruction inherently more challenging than single-step accuracy.

Fig. 2c compares the top-\textit{k} pathway accuracy of FlowER and G2S, showing the impact of their different approaches to predicting reaction sequences.

This discrepancy between top-1 step accuracy and top-1 pathway accuracy can be attributed to differences in how FlowER and G2S distribute predictions across reaction sequences. 
In FlowER, top-1 predictions for individual steps are highly accurate, but non-top-1 predictions are evenly distributed across sequences. 
This pattern reflects FlowER's ability to explore diverse mechanistic possibilities for each elementary step. 
However, when evaluating complete pathways, these occasional non-top-1 predictions can accumulate, leading to a reduction in top-1 pathway accuracy.

In contrast, G2S tends to cluster non-top-1 predictions within specific sequences. 
While this limits the diversity of explored pathways, it results in fewer sequences being affected by prediction errors. 
As a result, G2S achieves higher top-1 pathway accuracy despite lower top-1 step accuracy. 
FlowER’s stronger performance in top-2 pathway accuracy highlights its broader exploration of alternative pathways, possibly providing valuable insights into reaction mechanisms that may not be captured by G2S.

\clearpage
\section{Hyperparameter tuning}\label{si_hyperparam}

\begin{table}[h!]
\centering
\caption{Hyperparameter setting used in the experiments for different datasets. Best settings selected based on validation are highlighted in \textbf{bold} if multiple values have been experimented.}
\label{tab:hyperparameters}
\begin{tabular}{@{}lllp{5cm}@{}}
\toprule
Model           & Dataset & Parameter                      & Value(s)         \\ \midrule
FlowER          &  All  & Embedding size                   & \textbf{128}, 256 \\
                &       & Hidden size (same among all modules) & \textbf{128}, 256 \\
                &       & Filter size in Transformer      & 2048            \\
                &       & Attention encoder layers       &  6, 8, \textbf{12},   \\
                &       & Attention encoder heads        & 8, \textbf{32}       \\
                &       & Sigma                         & 0.1, 0.13, \textbf{0.15}, 0.16, 0.17, 2.0       \\
                &       & RBF low, high, step          & \textbf{\{0, 8, 0.1\}}, \{0, 20, 0.1\}  \\
                &       & Learning rate                & \textbf{0.0001}, 0.001       \\ 
                &       & Sample size                  & 16, \textbf{32}       \\ 
                &       & Scheduler                    & \textbf{NoamLR}, StepLR, None       \\ 
                \bottomrule
FlowER          &  Full & Total number of steps        & 1500000       \\ 
                \bottomrule
FlowER          &  50k, 5k, 
                1500, 500 & Total number of steps        & 500000       \\ 
                \bottomrule
                \bottomrule

Graph2SMILES    &  All  & Embedding size                 & \textbf{256}, 512 \\
                &       & Hidden size (same among all modules) & \textbf{256}, 512 \\
                &       & Filter size in Transformer     & 2048            \\
                &        & Number of D-MPNN layers        & 2, \textbf{4}, 6 \\
                &       & Attention encoder layers       & 4, \textbf{6}   \\
                &       & Attention encoder heads        & 8              \\
                &       & Decoder layers                 & 4, \textbf{6}   \\
                &       & Decoder heads                  & 8              \\
                &       & Number of accumulation steps   & 4              \\ 
                &       & Batch size                     & 4096            \\ 
                \bottomrule
Graph2SMILES    &  Full & Total Steps                    & 1200000          \\ 
                &       & Noam learning rate factor      & 2          \\ 
                &       & Dropout                        & 0.1          \\ 
                \bottomrule
Graph2SMILES    &  50k, 5k, 
                1500, 500 & Total Steps                  & 300000          \\ 
                &       & Noam learning rate factor      & 2, \textbf{4}          \\ 
                &       & Dropout                        & 0.1, \textbf{0.3}      \\ 
                \bottomrule
\end{tabular}
\end{table}

\clearpage
\section{Low data regime experiments}\label{si_lowdata}
A data-constrained evaluation is carried out to demonstrate the effectiveness of FlowER as compared to sequence translation based approach in predicting reaction outcome with only very small amount of data. Training data is randomly subsampled into smaller partitions of 50k, 5k, 1500, and 500. Using a training set size of 1500 roughly corresponds to 1 example per reaction class \& condition pair from the full training set. Despite this very small amount of data, even when training from scratch, FlowER's problem formulation and architecture provides impressive evidence of generalization on the reaction outcome prediction task.

\clearpage
\section{Recovering reaction pathways with FlowER}

We analyzed reactions from patents reported in 2024 \cite{pistachio} that were not assigned to any of the 2,639 predefined reaction types in NameRxn \cite{namerxn}. These cases include reactions that may involve unexpected product formations, combinations of known transformations, or mechanistic pathways that are not explicitly categorized within existing classification schemes.

We performed a narrow beam search (width 2, depth 9) on 22,000 unrecognized reactions and successfully recovered 351 products. 

\begin{figure*}[h]
\centering
\includegraphics[width=\textwidth]{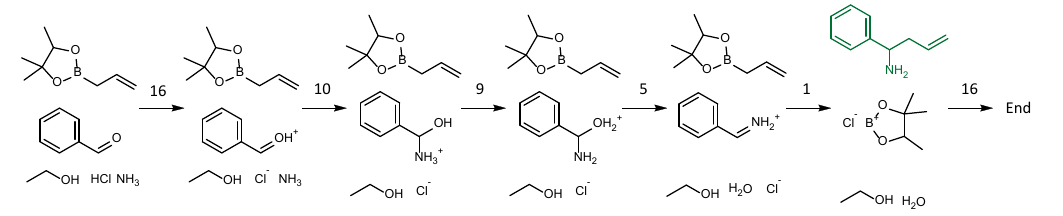}
\caption{
An example reaction reported in 
2024 \cite{US20240124445A1}, which was not assigned to a specific reaction class in our dataset \cite{pistachio, namerxn}.
FlowER successfully reproduces the experimentally-recorded product in green. Noteworthy in this example is the proper handling of intermediate protonation states throughout the acid-mediated amination (i.e., formation of an oxonium prior to nucleophilic attack); formation of the iminium in strongly acidic conditions is well-precedented \cite{afanasyev2019reductive}. Direct $\alpha$-allylation of imines has been described by \citet{alam2014stereoselective}; a Zimmerman-Traxler six-membered transition state is proposed \cite{mejuch2013axial}. The numbers above each arrow represent the number of times that reaction was proposed during 16 independent sampling steps. }\label{fig_unrecognized_8}
\end{figure*}

\begin{figure*}[h]
\centering
\includegraphics[width=\textwidth]{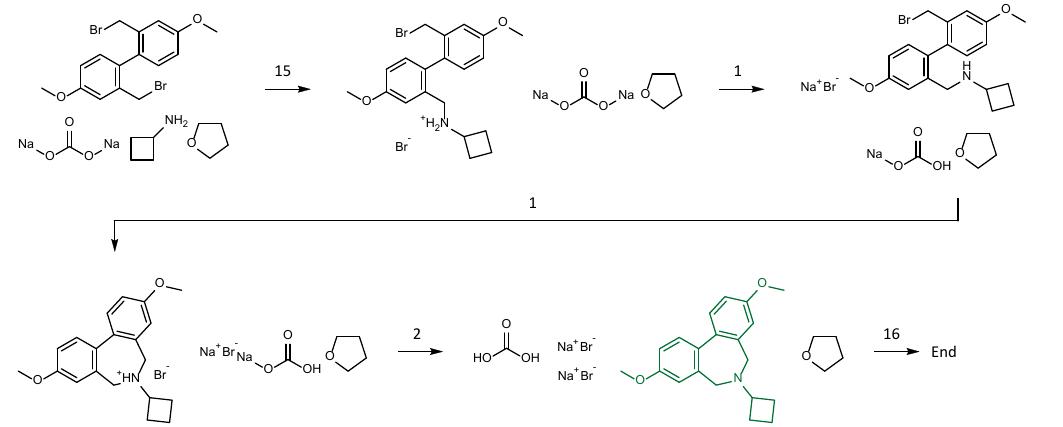}
\caption{
An example reaction reported in 
2024 \cite{US20240139149A1}, which was not assigned to a specific reaction class in our dataset \cite{pistachio, namerxn}. 
FlowER successfully reproduces the experimentally-recorded product in green. FlowER demonstrates strong  performance when handling the first \pKa-dependent substitution (ammonium \pKa = $\sim$9-11, carbonate \pKa = $\sim$10.33). However, the second deprotonation (en route to dihydro-dibenzo[\textit{c,e}]azepine) requires a second equivalent of carbonate to proceed (bicarbonate \pKa = $\sim$ 6.35). While FlowER has not seen examples of bicarbonate-mediated deprotonations, the reaction was not recorded with the required number of carbonates; accordingly, FlowER uses the 'best available' base to provide a reasonable pathway to the observed product despite this omission. This example showcases the need for detailed procedural record-keeping to build next-generation models that are mass-balance- and mechanism-aware.  The numbers above each arrow represent the number of times that reaction was proposed during 16 independent sampling steps. }\label{fig_unrecognized_1}
\end{figure*}

The five reaction pathways successfully recovered by FlowER, originating from reactions reported in nine patents from 2024 \cite{US20240139149A1, US20240101584A1, US20240116946A1, US20240239807A1, US20240128510A1, US20240150295, US20240150296, US20240140962A1, US20240122180A1, US20240124445A1, US20240118617A1}, are illustrated in Fig.~\ref{fig_unrecognized_8} through Fig.~\ref{fig_unrecognized_9}.

It is important to note that the reactants shown here correspond to those recorded in Pistachio \cite{pistachio}, including both reactants and reagents. However, certain chemicals listed in the patents may be missing, or additional compounds introduced during work-up might be included. Furthermore, the overall neutralization reaction was not explicitly considered when curating the mechanistic dataset, meaning that some molecules may remain unneutralized. Additionally, since the stronger acid/base species were not prioritized for use in FlowER’s design, the model may not predict these cases accurately.

\begin{figure*}[h]
\centering
\includegraphics[width=\textwidth]{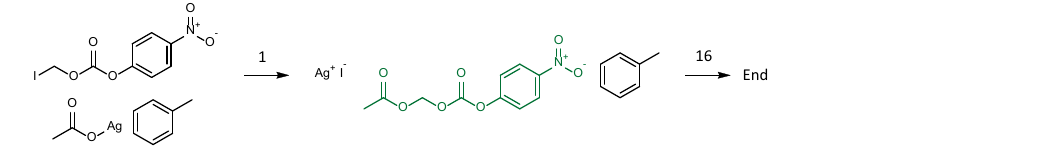}
\caption{
An example reaction reported in 
2024 \cite{US20240101584A1}, which was not assigned to a specific reaction class in our dataset \cite{pistachio, namerxn}. 
FlowER successfully reproduces the experimentally-recorded product in green. Despite the operational simplicity of the above mechanism, FlowER was provided no examples at training incorporating Ag. Successful reconstruction of this S$_\text{N}$2 mechanism demonstrates a surprising ability to handle out-of-distribution elements, correctly generating the expected ester. The numbers above each arrow represent the number of times that reaction was proposed during 16 independent sampling steps. }\label{fig_unrecognized_2}
\end{figure*}

\begin{figure*}[h]
\centering
\includegraphics[width=\textwidth]{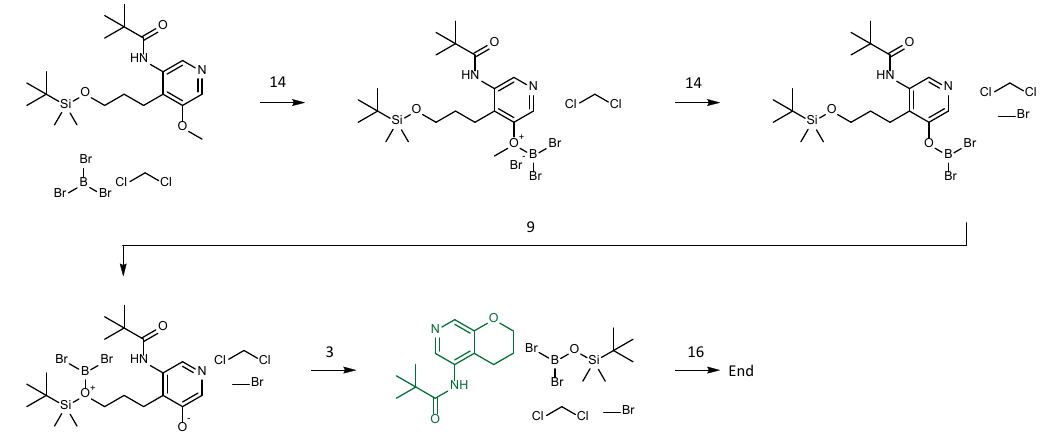}
\caption{
An example reaction reported in 
2024 \cite{US20240116946A1, US20240239807A1}. 
FlowER successfully reproduces the experimentally-recorded product in green. Importantly, FlowER captures the commonly accepted unimolecular mechanism for demethylation \cite{benton1942cleavage}, formalized by \citet{mcomie1968demethylation} in 1968. Closer inspection of the patent conditions reveals the ground truth conditions use of 3 equivalents of \ce{BBr3}. This offers a potential alternative mechanism, wherein the \textit{tert}-butyldimethylsilyl (TBS) ether activation is accomplished with a second equivalent of \ce{BBr3}. We note this nuanced mechanism is still actively being researched \cite{kosak2015ether}, and several alternative mechanisms have been proposed \cite{sousa2013bbr3}. The numbers above each arrow represent the number of times that reaction was proposed during 16 independent sampling steps.}\label{fig_unrecognized_2}
\end{figure*}

\begin{figure*}[h]
\centering
\includegraphics[width=\textwidth]{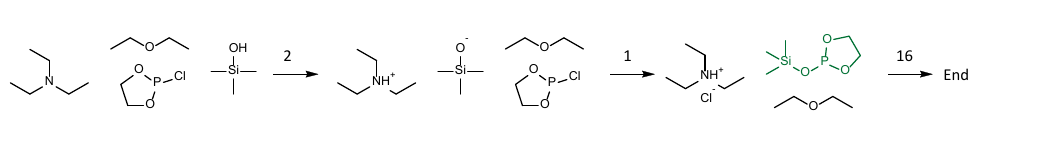}
\caption{
An example reaction reported in 
2024 \cite{US20240128510A1}. 
FlowER successfully reproduces the experimentally-recorded product in green. We select this particular example to highlight FlowER's handling of a substitution reaction at a non-carbon atom (i.e., the phosphorus of a phosphochloridite). Interestingly, FlowER opts for an anionic mechanism with the trimethylhydroxysilane (\pKa = $\sim$ 11) in the presence of TEA (10.8). This suggests the neutral hydroxysilane would operate as the primary nucleophile under these conditions,  though the similar relative acidities of both substrates provide an avenue for both mechanistic pathways. The numbers above each arrow represent the number of times that reaction was proposed during 16 independent sampling steps.}\label{fig_unrecognized_4}
\end{figure*}

\begin{figure*}[h]
\centering
\includegraphics[width=\textwidth]{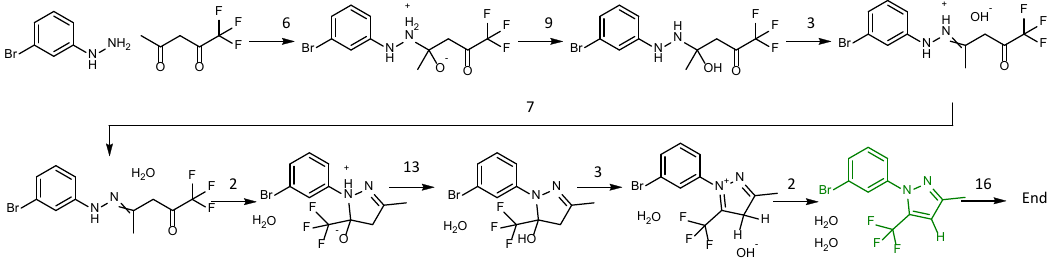}
\caption{
An example reaction reported in 
2024 \cite{US20240150295, US20240150296}. 
FlowER successfully reproduces the experimentally-recorded product in green. We highlight this example to showcase a named reaction that was not identified in Pistachio due to two recorded main products (See the Extended Fig. S2a); Specifically, the acid-mediated Knorr pyrazole synthesis \cite{knorr1883einwirkung}. While the commonly accepted mechanism in strongly acidic solution begins with formation of an oxonoium intermediate \cite{flood2018leveraging}, we find FlowER exhibits a preference for formation of a Zwitterionic intermediate. Nonetheless, FlowER accurately generates the required oxonium en route to both dehydration steps. This is noteworthy, as the formation of hydroxide in strongly acidic conditions can be problematic for other models. Additionally, FlowER accurately predicts the formation of both possible pyrazoles, showcasing the utility of FlowER for exploring mechanistic pathways leading to multiple possible products. The numbers above each arrow represent the number of times that reaction was proposed during 16 independent sampling steps. }\label{fig_unrecognized_5}
\end{figure*}

\begin{figure*}[h]
\centering
\includegraphics[width=\textwidth]{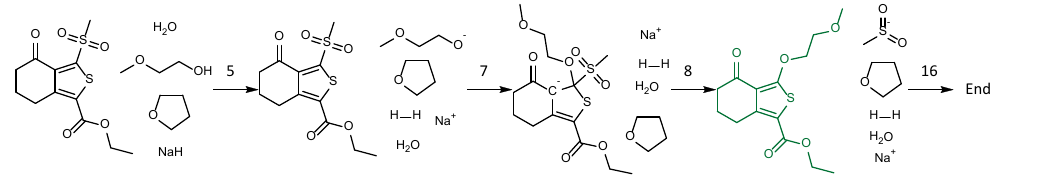}
\caption{
An example reaction reported in 
2024 \cite{US20240140962A1}. 
FlowER successfully reproduces the experimentally-recorded product in green. In this example, we showcase the ability for FlowER to identify and apply named reaction mechanisms (i.e., sulfone-mediated S$_\text{N}$Ar reaction \cite{patel2020sulfone}). We highlight the correct handling of terminal alcohols in the presence of \ce{NaH}, and the subsequent release of methanesulfinate to prepare the final fused (2-methoxyethoxy)thiophene. The numbers above each arrow represent the number of times that reaction was proposed during 16 independent sampling steps. }\label{fig_unrecognized_6}
\end{figure*}

\begin{figure*}[h]
\centering
\includegraphics[width=\textwidth]{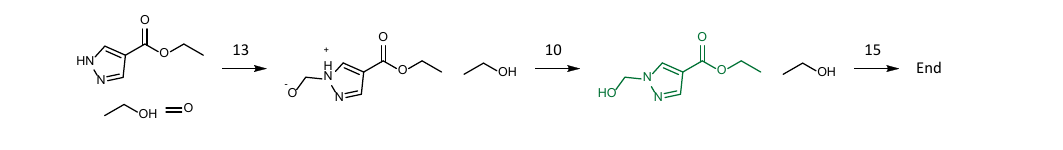}
\caption{
An example reaction reported in 
2024 \cite{US20240122180A1}.
FlowER successfully reproduces the experimentally-recorded product in green. Importantly, without the presence of an exogenous base, imidazole is correctly proposed to form a Zwitterionic intermediate upon nucleophilic attack of formaldehyde (rather than first undergoing deprotonation). The numbers above each arrow represent the number of times that reaction was proposed during 16 independent sampling steps. }\label{fig_unrecognized_7}
\end{figure*}

\begin{figure*}[h]
\centering
\includegraphics[width=\textwidth]{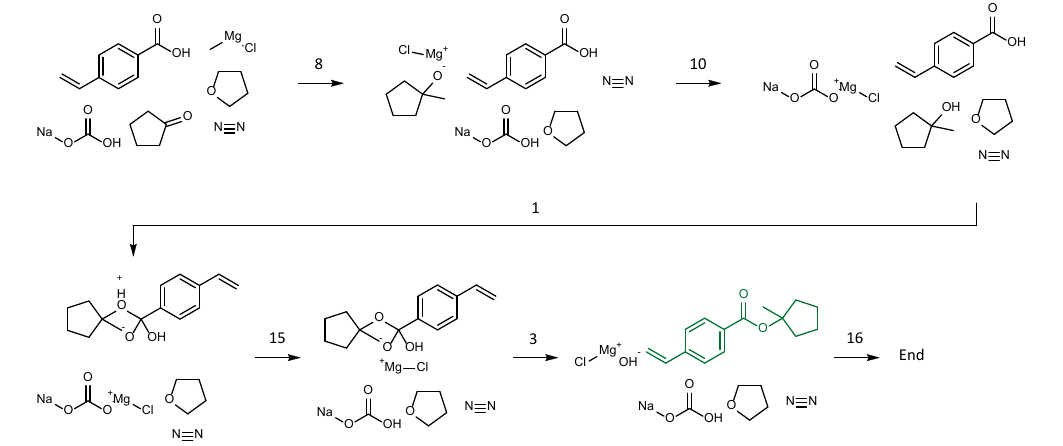}
\caption{
An example reaction reported in 
2024 \cite{US20240118617A1}. 
FlowER successfully reproduces the experimentally-recorded product in green. 
We highlight this example to showcase how FlowER navigates the entire complex reagent pool to identify the correct mechanism. 
While order of addition is undoubtedly of utmost importance in this reaction (i.e., Grignard reagents will rapidly act as a strong base in the presence of a carboxylic acid), FlowER is able to identify a reasonable mechanistic pathway to the observed product. In this particular case, FlowER implicitly identifies the requirement for two subsequent reactions: a Grignard addition and base-mediated esterification. The numbers above each arrow represent the number of times that reaction was proposed during 16 independent sampling steps. }\label{fig_unrecognized_9}
\end{figure*}

\clearpage

\clearpage
\section{Fine-tuning on previously-unseen reaction types}
\subsection{Performance on out-of-distribution reaction types}
To evaluate FlowER’s ability to generalize to new reaction types, we selected 12 reaction types that were not included in the original training set: Transamidation, Prilezhaev epoxidation, Diels-Alder cycloaddition, Staudinger reduction, Hydroxy to bromo, N-Bn deprotection, Negishi coupling, Carboxylic acid sulfonamide condensation, Menshutkin reaction, Urea Schotten-Baumann, Appel bromination, and SEM protection. These reaction types were curated using the same method for curating our mechanistic dataset as described in Section \ref{si_dataset}.  

For fine-tuning, only 32 overall reactions were used as the training set. The validation set contained at least two overall reactions per reaction type, while the remaining reactions were assigned to the test set. 

\begin{figure*}[h]
\centering
\includegraphics[width=0.95\textwidth]{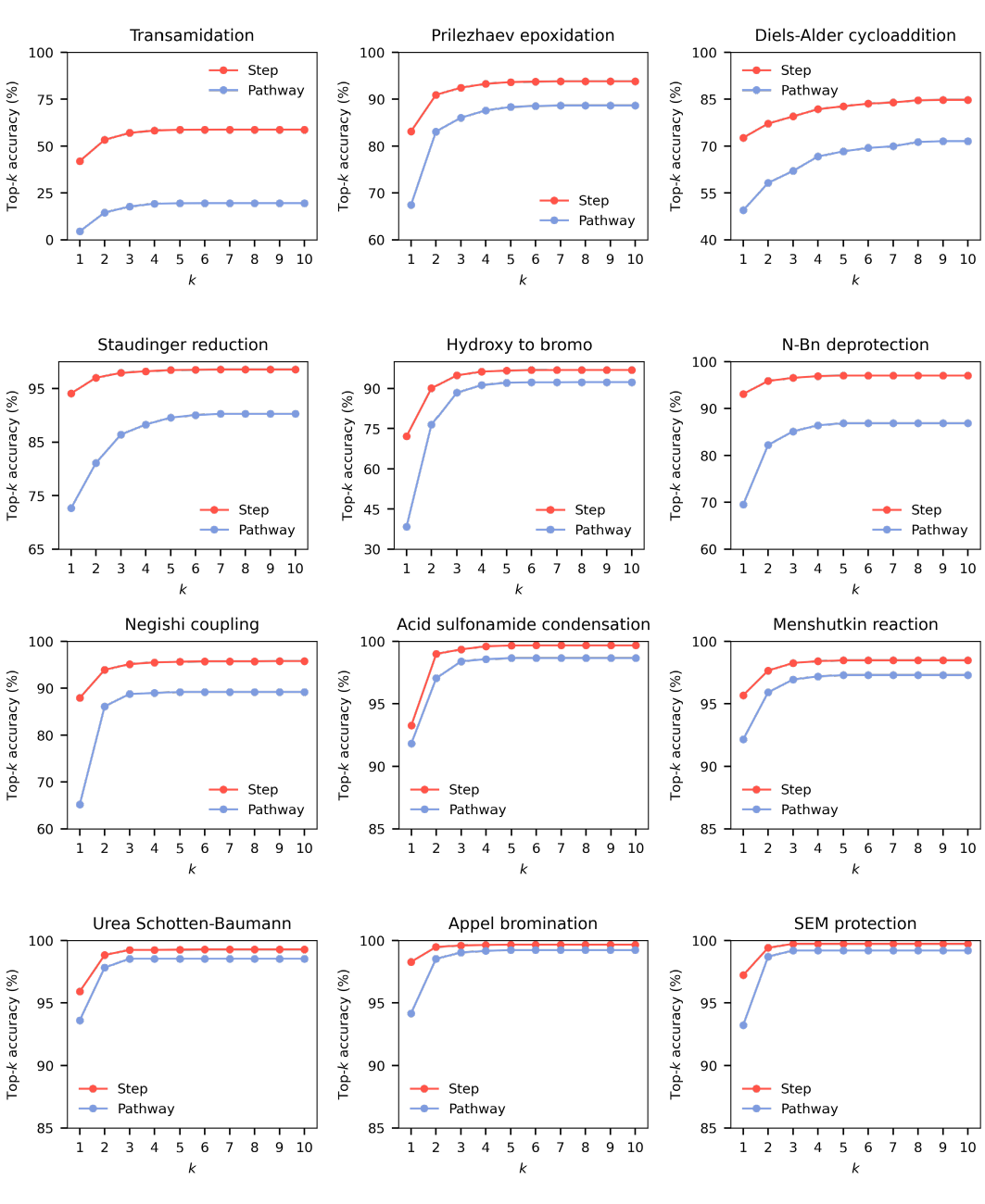}
\caption{
Top-\textit{k} accuracy of FlowER on previously unseen reaction types after fine-tuning on 32 examples. 
Each subplot represents the performance on one of the 12 reaction classes that were excluded from the initial training set. 
Red and blue curves indicate top-\textit{k} step accuracy and top-\textit{k} pathway accuracy, respectively. 
}\label{fig_fine_tune_performance}
\end{figure*}

The best-performing model, selected based on validation set accuracy, was evaluated on the test set. The top-\textit{k} step and pathway accuracies on the test set are presented in Fig.~\ref{fig_fine_tune_performance}.

The elementary steps of the 12 reaction types and their top-1 step accuracy before and after fine-tuning can be found in Fig.~\ref{fig_ood_1,2,3} through Fig.~\ref{fig_ood_12}.

\subsection{Assessing catastrophic forgetting after fine-tuning}

To investigate whether fine-tuning caused the model to catastrophically forget reactions learned during the original training phase, we evaluated FlowER and its fine-tuned variants on a subset of the original test set. Specifically, we randomly sampled 10\% of the original test set and compared the performance of the original FlowER model with the fine-tuned models.  

\begin{figure*}[h]
\centering
\includegraphics[width=0.95\textwidth]{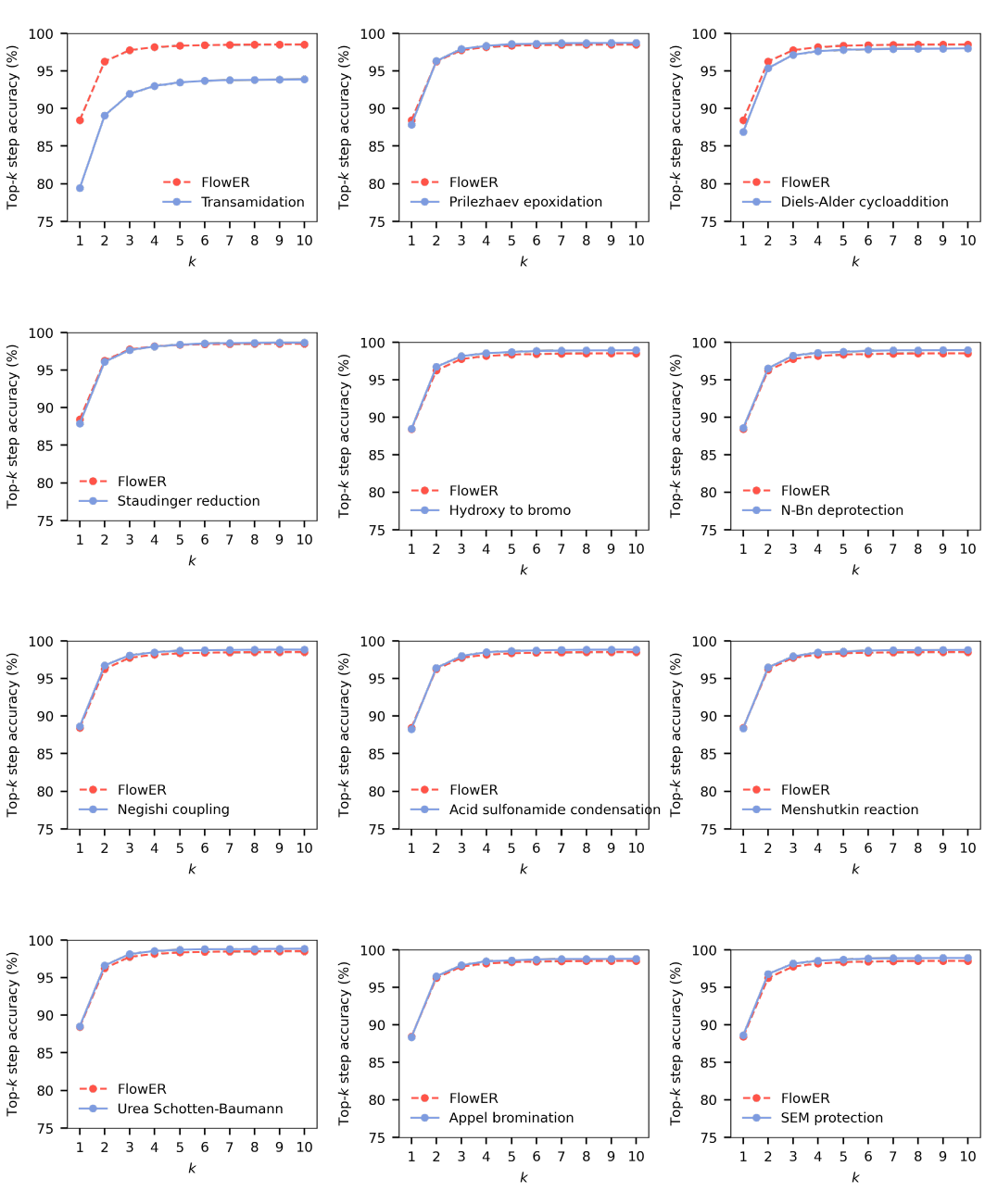}
\caption{
Performance of fine-tuned models on a random 10\% subset of the original test set compared to the original FlowER without fine-tuning, in terms of top-$k$ step accuracy. 
This evaluation was conducted to assess catastrophic forgetting, where a model forgets previously learned knowledge after fine-tuning on new reaction types. 
For each fine-tuned model, we recomputed the top-$k$ step accuracy on the sampled test subset and compared it against the original FlowER model.
}\label{fig_step_forgetting}
\end{figure*}

\begin{figure*}[h]
\centering
\includegraphics[width=0.95\textwidth]{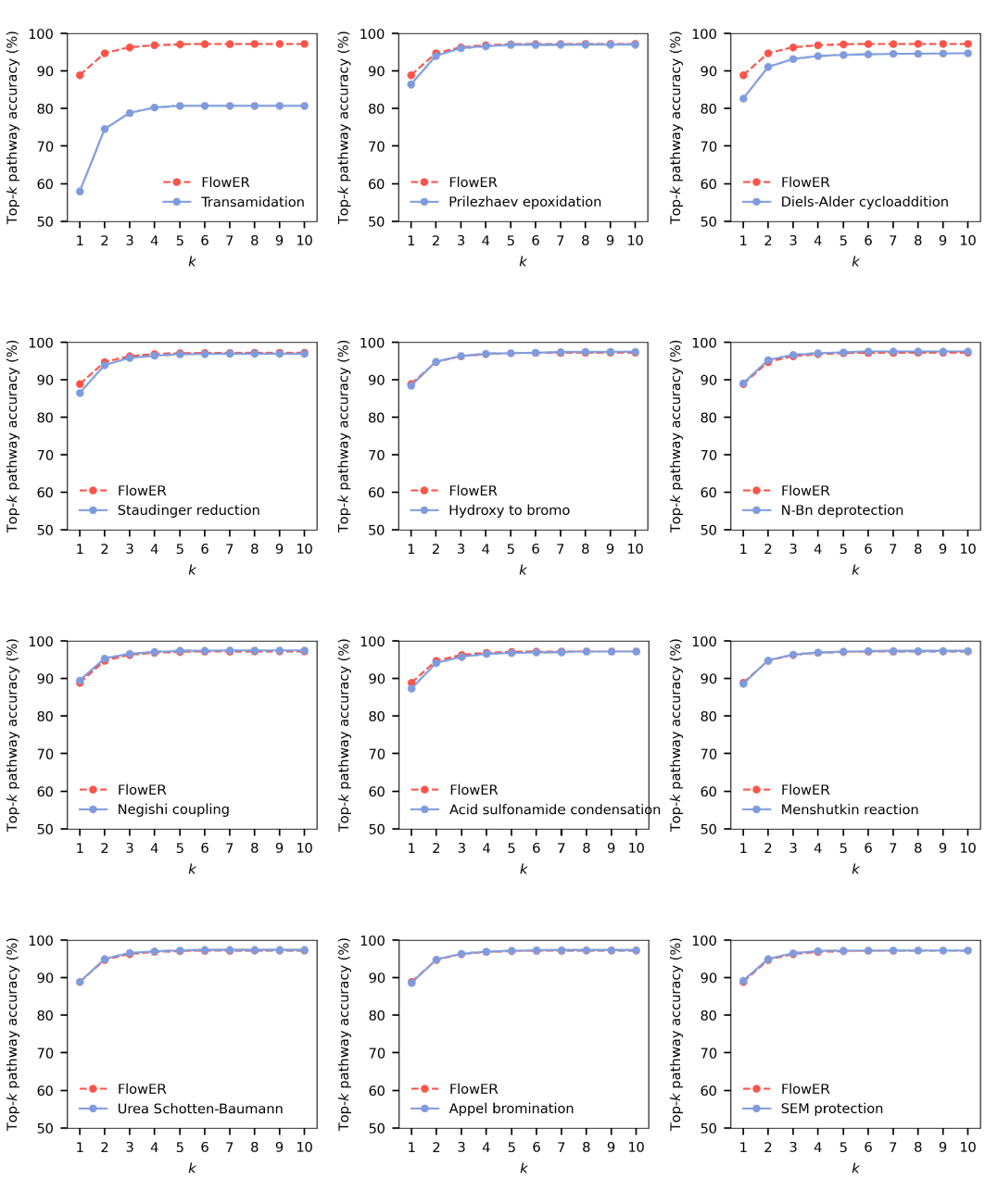}
\caption{
Performance of fine-tuned models on a random 10\% subset of the original test set compared to the original FlowER without fine-tuning, in terms of top-$k$ pathway accuracy. 
This evaluation was conducted to assess catastrophic forgetting, where a model forgets previously learned knowledge after fine-tuning on new reaction types. 
For each fine-tuned model, we recomputed the top-$k$ pathway accuracy on the sampled test subset and compared it against the original FlowER model.
}\label{fig_path_forgetting}
\end{figure*}

The top-\textit{k} step accuracy for FlowER and the fine-tuned models is presented in Fig.~\ref{fig_step_forgetting}, while the top-\textit{k} pathway accuracy is shown in Fig.~\ref{fig_path_forgetting}. 

Among the 12 fine-tuned models, 11 showed no substantial drop in accuracy compared to the original model, indicating minimal catastrophic forgetting. The exception was the Transamidation fine-tuned model, which exhibited a noticeable performance drop. This suggests that, in most cases, FlowER retains its ability to predict previously learned reactions even after fine-tuning on unseen reaction types.

\clearpage

\clearpage

\clearpage

\begin{figure*}[h]
\centering
\includegraphics[width=0.95\textwidth]{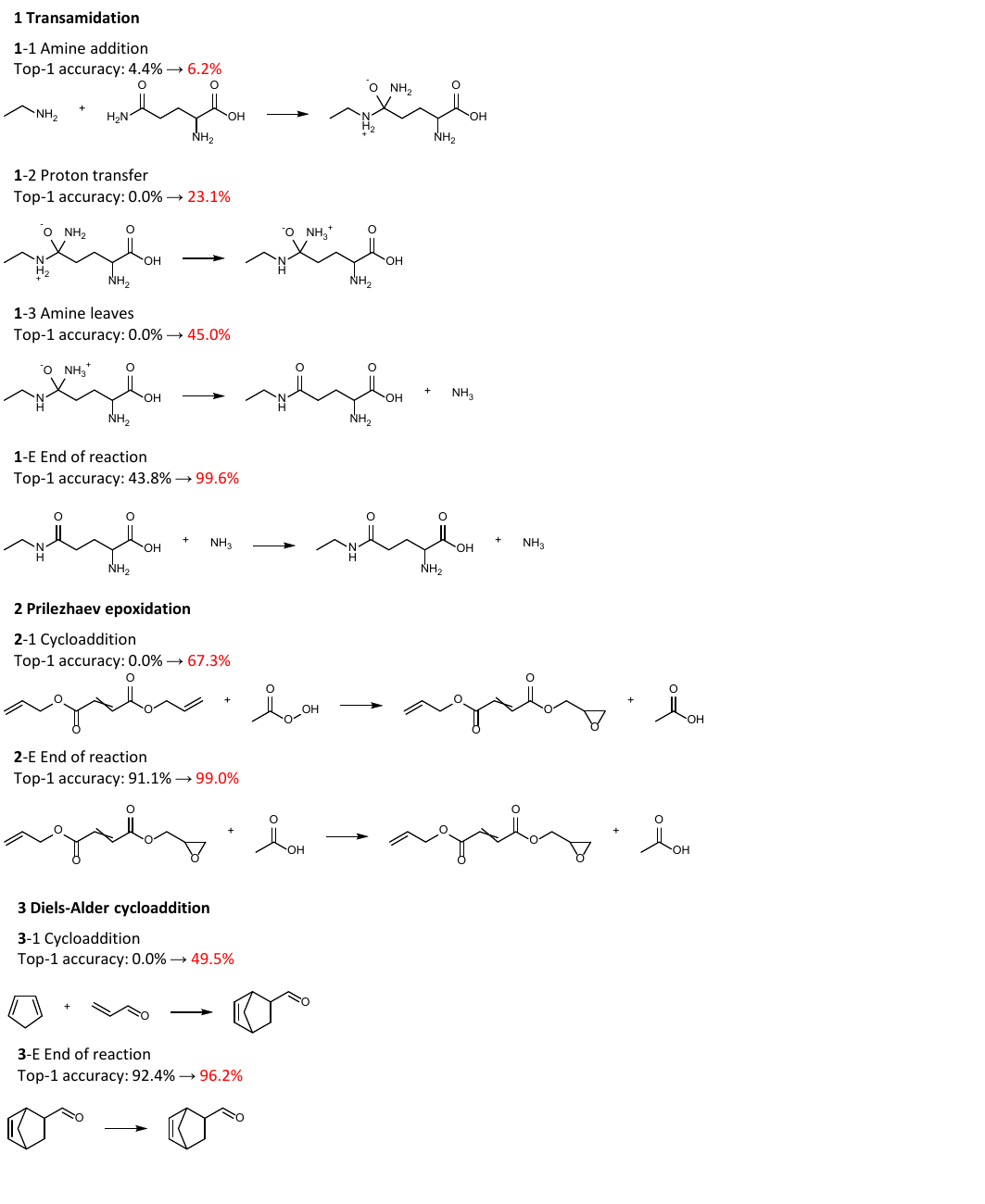}
\caption{
Changes in top-1 step accuracy for elementary steps of \textbf{1} transamidation, \textbf{2} Prilezhaev epoxidation, and \textbf{3} Diels-Alder cycloaddition before and after fine-tuning on 32 reaction examples.
}\label{fig_ood_1,2,3}
\end{figure*}

\begin{figure*}[h]
\centering
\includegraphics[width=0.95\textwidth]{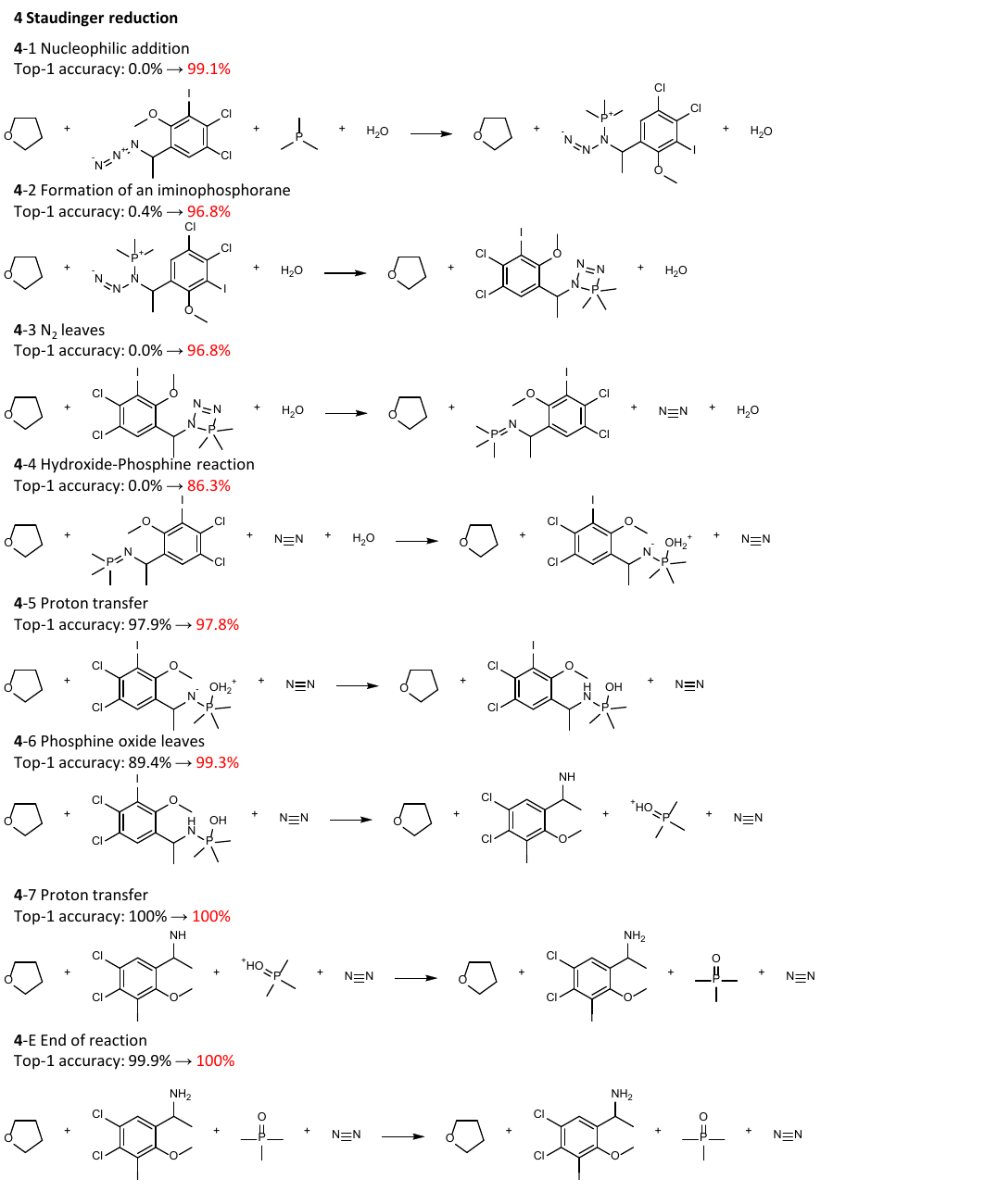}
\caption{
Changes in top-1 step accuracy for elementary steps of \textbf{4} Staudinger reduction before and after fine-tuning on 32 examples.
}\label{fig_ood_4}
\end{figure*}

\begin{figure*}[h]
\centering
\includegraphics[width=0.95\textwidth]{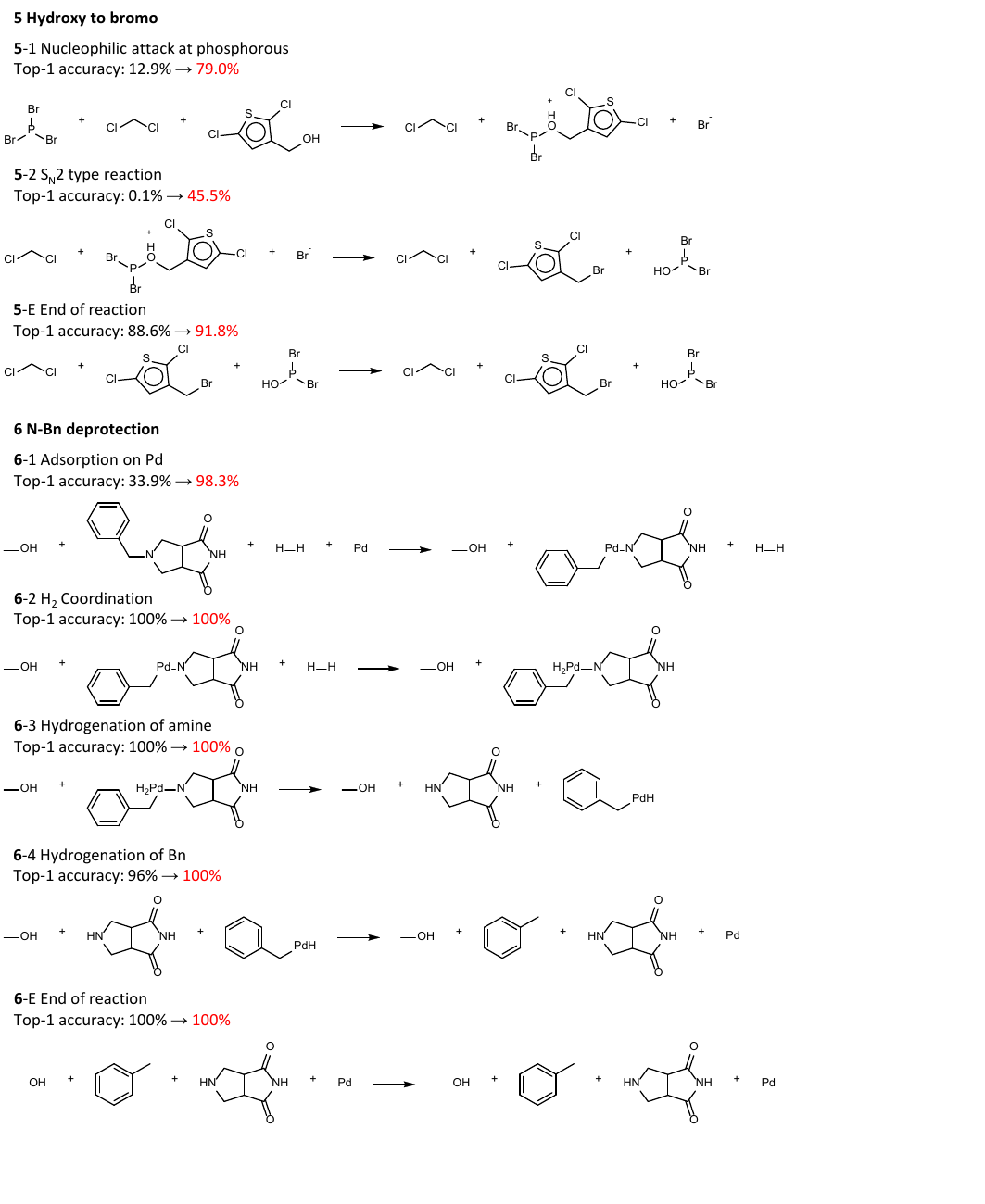}
\caption{
Changes in top-1 step accuracy for elementary steps of \textbf{5} Hydroxy to bromo and \textbf{6} N-Bn deprotection before and after fine-tuning on 32 examples.
}\label{fig_ood_5,6}
\end{figure*}

\begin{figure*}[h]
\centering
\includegraphics[width=0.95\textwidth]{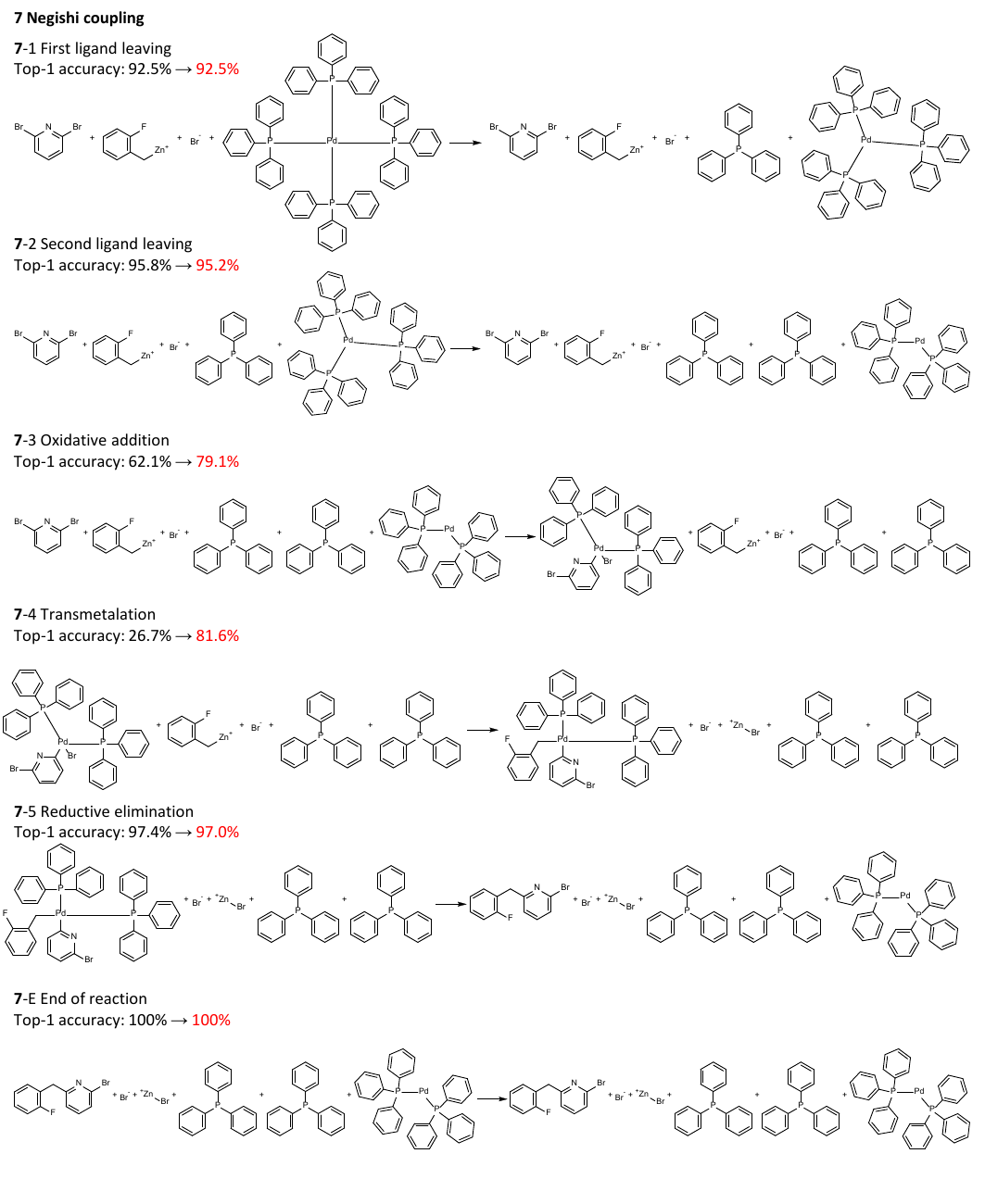}
\caption{
Changes in top-1 step accuracy for elementary steps of \textbf{7} Negishi coupling before and after fine-tuning on 32 examples.
}\label{fig_ood_7}
\end{figure*}

\begin{figure*}[h]
\centering
\includegraphics[width=0.95\textwidth]{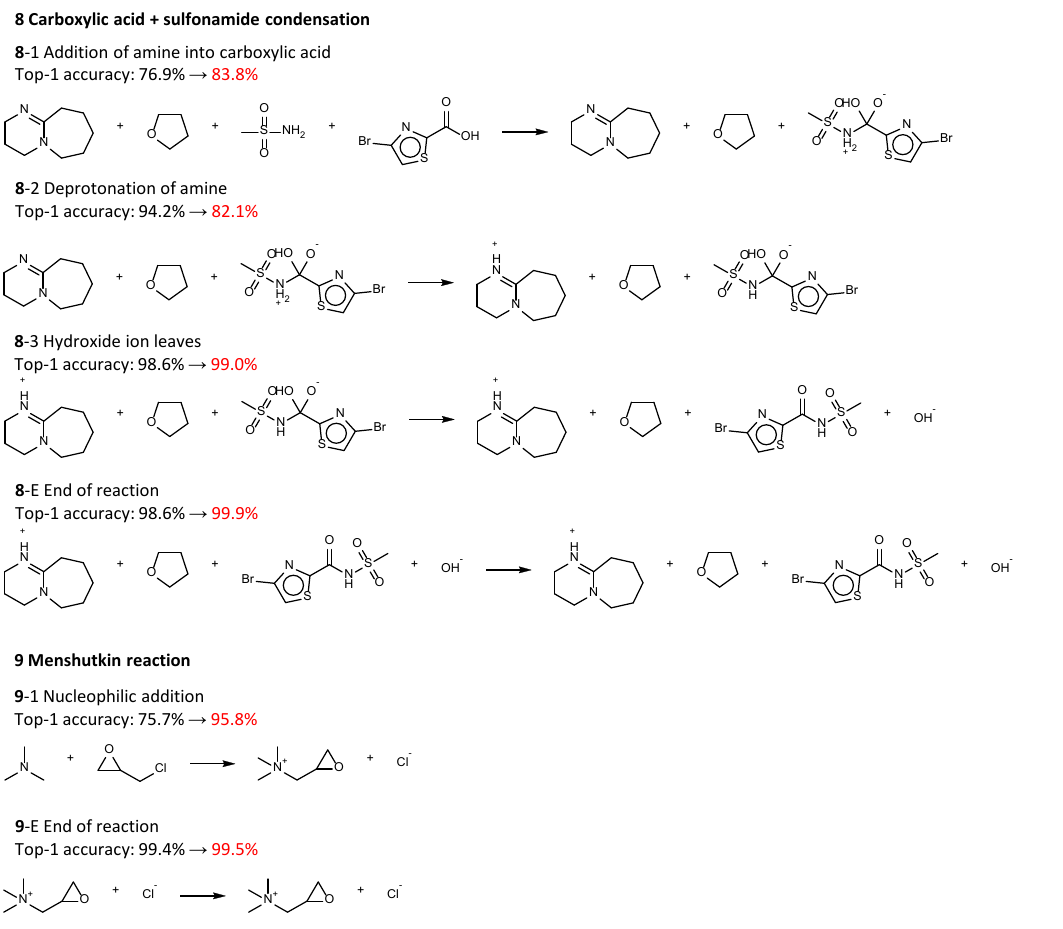}
\caption{
Changes in top-1 step accuracy for elementary steps of \textbf{8} Carboxylic acid + sulfonamide condensation and \textbf{9} Menshutkin reaction before and after fine-tuning on 32 examples.
}\label{fig_ood_8,9}
\end{figure*}

\begin{figure*}[h]
\centering
\includegraphics[width=0.95\textwidth]{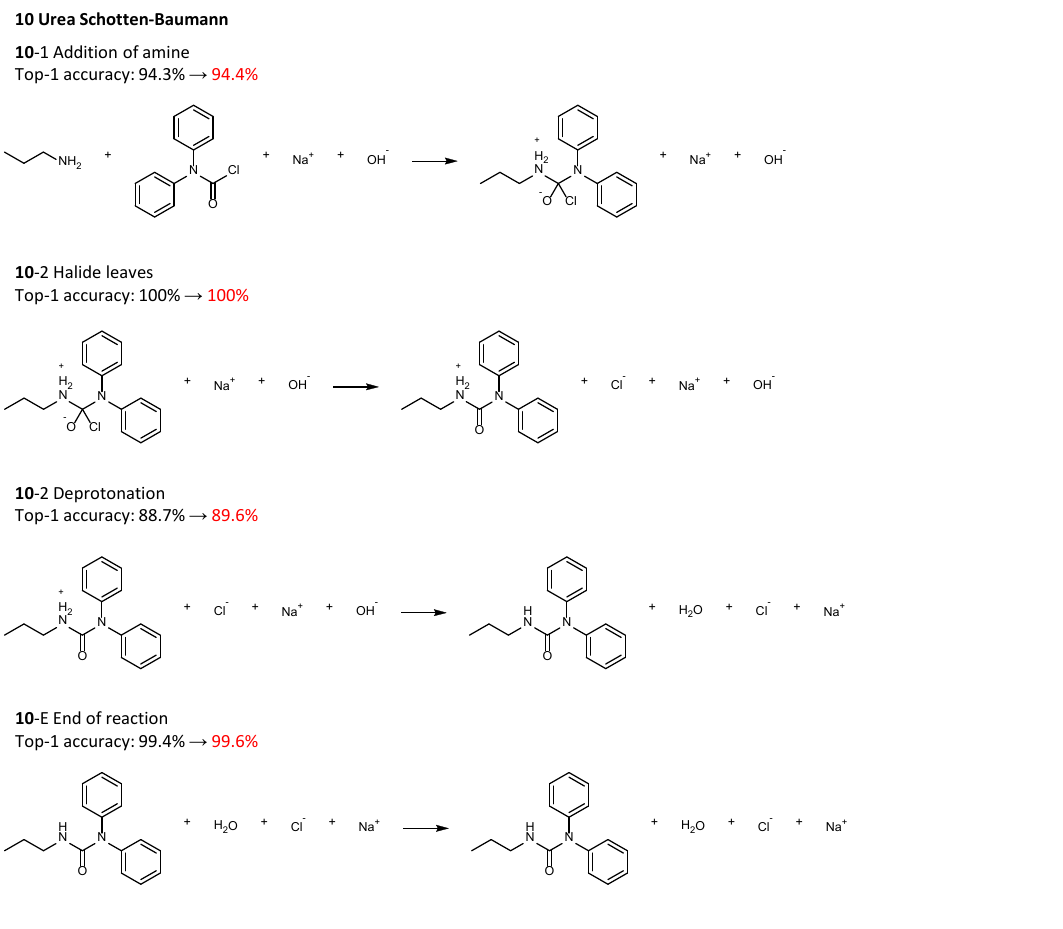}
\caption{
Changes in top-1 step accuracy for elementary steps of \textbf{10} Urea Schotten-Baumann before and after fine-tuning on 32 examples.
}\label{fig_ood_10}
\end{figure*}

\begin{figure*}[h]
\centering
\includegraphics[width=0.95\textwidth]{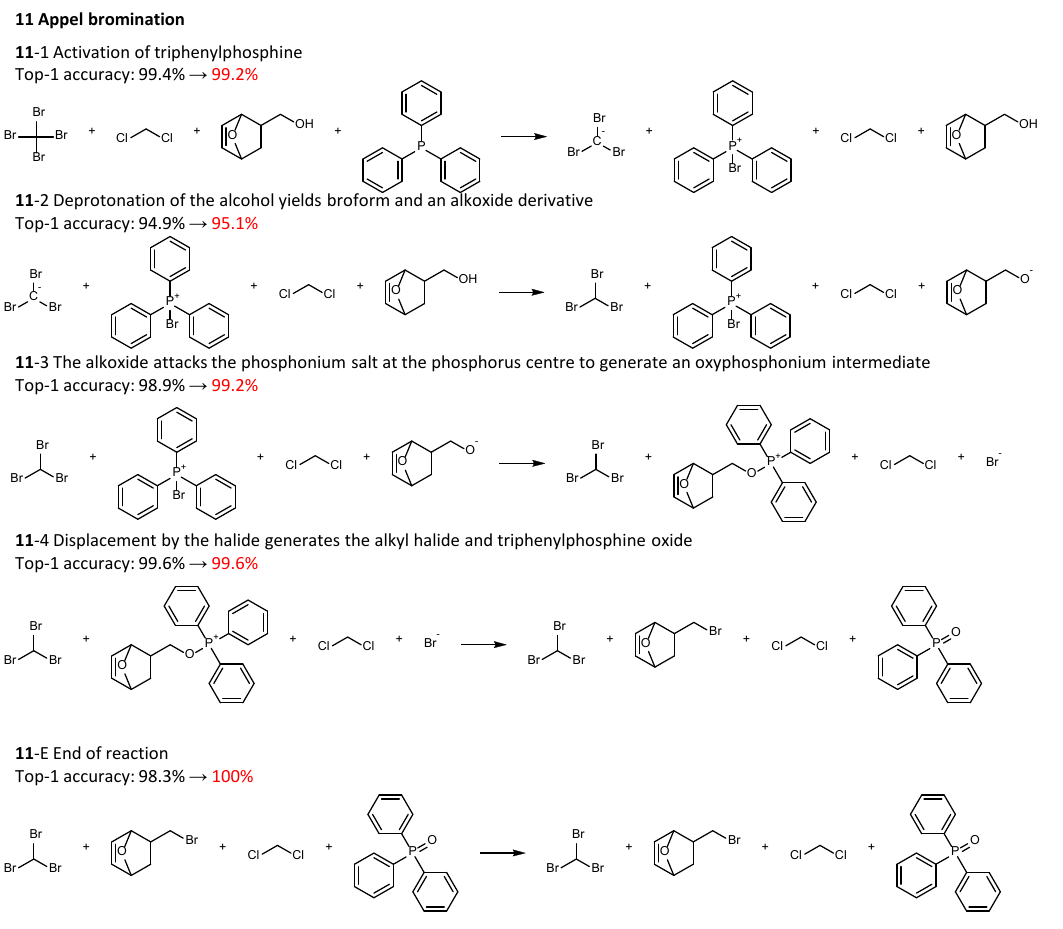}
\caption{
Changes in top-1 step accuracy for elementary steps of \textbf{11} Appel bromination before and after fine-tuning on 32 examples.
}\label{fig_ood_11}
\end{figure*}

\begin{figure*}[h]
\centering
\includegraphics[width=0.95\textwidth]{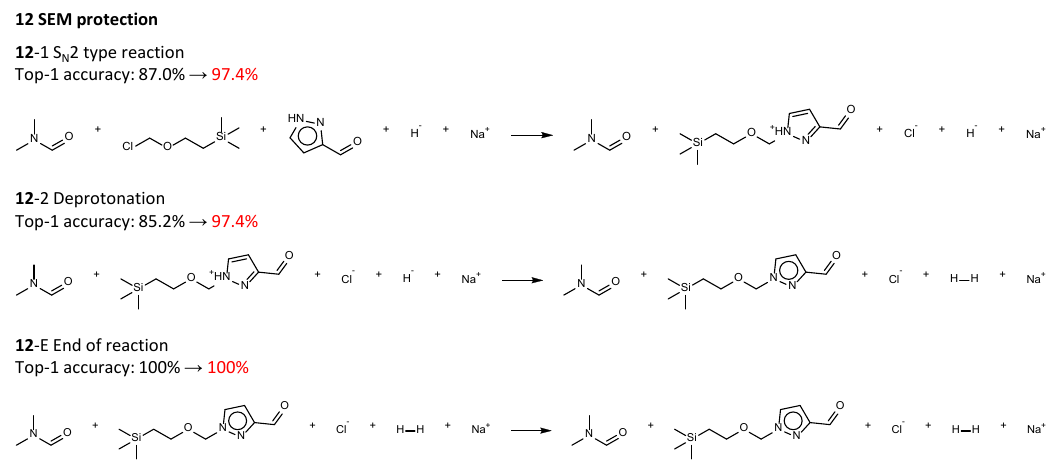}
\caption{
Changes in top-1 step accuracy for elementary steps of \textbf{12} SEM protection before and after fine-tuning on 32 examples.
}\label{fig_ood_12}
\end{figure*}

\clearpage

\bibliography{sn-bibliography,references}